\documentclass[twocolumn]{autart}    

\usepackage{amsmath,amssymb,amsfonts}
\usepackage{graphicx}
\usepackage[colorlinks=true, allcolors=blue]{hyperref}
\usepackage{soul}
\usepackage{booktabs} 
\usepackage{multirow} 

\newcommand{\z}{{\bf z}}

\newcommand{\m}{{\bf m}}
\newcommand{\uu}{{\bf u}}

\newcommand{\y}{{\bf y}}

\newcommand{\btheta}{\boldsymbol{\theta}}

\newtheorem{example}{Example}
\newtheorem{lemma}{Lemma}
\newtheorem{proposition}{Proposition}
\newtheorem{remark}{Remark}
\usepackage{breqn}

\begin{document}

\begin{frontmatter}

\title{dynoGP: Deep Gaussian Processes for dynamic system identification\thanksref{footnoteinfo}} 
\thanks[footnoteinfo]{This paper was not presented at any IFAC conference.}
\thanks[footnoteinfo2]{This work was partly  supported by the Eurostars programme, project``E3093 - SLIMPEC: A Software suite for LearnIng-based embedded Model PrEdictive Control''.}

\author[Dublin]{Alessio Benavoli}\ead{alessio.benavoli@tcd.ie}, 
\author[Lugano-Viganello]{Dario Piga}\ead{dario.piga@supsi.ch}, 
\author[Lugano-Viganello]{Marco Forgione}\ead{marco.forgione@supsi.ch},   
\author[Lugano-Viganello]{Marco Zaffalon}\ead{marco.zaffalon@supsi.ch}  

\address[Dublin]{School of Computer Science and Statistic, Trinity College Dublin,  Ireland}   
\address[Lugano-Viganello]{SUPSI, IDSIA - Dalle Molle Institute for Artificial Intelligence, Lugano, Switzerland}   

\begin{abstract}                          
In this work, we present a novel approach to system identification for dynamical systems, based on a specific class of Deep Gaussian Processes (Deep GPs). These models are constructed by interconnecting linear dynamic GPs (equivalent to stochastic linear time-invariant dynamical systems) and static GPs (to model static nonlinearities). Our approach combines the strengths of data-driven methods, such as those based on neural network architectures, with the ability to output a probability distribution. This offers a more comprehensive framework for system identification that includes uncertainty quantification. 
Using both simulated and real-world data, we demonstrate the effectiveness of the proposed approach.
\end{abstract}

\end{frontmatter}

\section{Introduction}

The application of neural networks to system identification dates back to the early 1990s, with seminal works by Chen et al. \cite{chen1990non}, Wray and Green \cite{wray1994neural}, and Narendra and Parthasarathy \cite{narendra1990identification}. These studies proposed the use of neural networks for black-box data-driven modeling and control of nonlinear dynamic systems, laying the foundation for future research in this area. The field has gained significant momentum in the last decade, with several notable advancements (see the survey paper \cite{pillonetto2025deep} for a comprehensive discussion on the topic). Among available studies, we mention the 1-D convolutional networks to describe Nonlinear Finite Impulse Response (NFIR) models \cite{andersson2019deep,bai2018empirical}, and the application of Recurrent Neural Networks (RNNs) and Long Short-Term Memory (LSTM) networks for efficiently describing long-term dynamics \cite{wang2017system}. Novel architectures, tailored to dynamical systems, have also emerged in recent years, such as: the autoencoder-based architectures \cite{beintema2021nonlinear,masti2021learning} which map input-output sequences into state variables; deep state-space models which preserve the interpretability of the networks and leverage, when available, prior physical knowledge \cite{gedon2021deepssm,forgione2020model}; and dynoNet \cite{forgione2021dynonet}, composed of the interconnection of Linear Time-Invariant (LTI) blocks described by transfer functions followed by static nonlinearities (activation functions or multi-layer perceptron). Similar to dynoNet, other architectures like Structured State Space Sequence models (S4) \cite{gu2022parameterization} and Linear Recurrent Units (LRUs) \cite{orvieto2023resurrecting} have been proposed, made by connection of linear dynamical blocks followed by static nonlinearities. Unlike dynoNet, the linear blocks are described in a state space form, with proper parametrisation of the matrices designed to ensure efficiency during training.

A defining characteristic of architectures that integrate linear dynamical blocks with static nonlinear blocks is their universal approximation capability. Indeed, any continuous nonlinear operator with fading memory can be approximated using a layer of linear dynamical blocks followed by a nonlinear (polynomial) static block; see for instance \cite{boyd1985fading,schetzen2006volterra}. 
The system identification methods mentioned above, such as dynoNet, share this universal approximation property; however, they only provide point estimates of the system's output and lack a component for characterising uncertainty, which is fundamental for reliable decision making (particularly in control applications).

 In contrast to deep neural networks, probabilistic models like deep Gaussian Processes (deepGPs) \cite{damianou2013deep} allow for the quantification of estimation uncertainty. These models combine the advantages of Gaussian Processes with the expressive power of deep neural networks, producing a probability distribution as output that can be leveraged for reliable decision-making. However,  deepGPs have not yet been developed for system identification.

The objective of this paper is to merge ideas underlying dynoNet (or equivalently, SS4 and LRU) with those of deepGP, aiming to develop dynamic deepGP models, which we call in this paper \emph{dynoGP}.
These models are constructed through the interconnection of dynamic GPs (specifically, as discussed in the rest of the paper, stochastic linear time-invariant dynamical systems) and static GPs. This approach seeks to leverage the strengths of both deterministic recursive neural network architectures  and probabilistic deep Gaussian Process models, potentially offering a more comprehensive framework for system identification that includes uncertainty quantification.

The rest of the paper is organised as follows. In Section \ref{sec:GP}, we provide an overview of GPs for regression, including a discussion on the selection of inducing points in sparse GP models, as this is essential for the dynoGP model proposed in this paper. In Section \ref{sec:LTI}, we introduce the concept of stochastic LTI dynamical systems with Brownian process noise, exploring the connections between LTI systems and GP. We review literature demonstrating that the output trajectory of such systems is  a Gaussian Process. We then present initial results, showing that for LTI systems with complex diagonal state-transition matrices, the mean and covariance functions of the GP describing the output trajectories can be simplified and computed in a closed form. In Section \ref{sec:dynoGP}, we introduce the deep dynamical GP architecture (dynoGP), constructed by stacking layers of dynamic and static GPs. We employ inducing points to approximate the posterior predictive distribution through stochastic variational inference~\cite{salimbeni2017doubly}. Section \ref{Sec:examples} presents four examples, three involving real datasets, to demonstrate the effectiveness of the proposed dynoGP architecture, alongside comparisons with other learning algorithms. Detailed mathematical derivations of the paper's results and specifics on implementing the dynoGP models are discussed in the Appendix.

\section{Gaussian Processes}
\label{sec:GP}
Gaussian Processes (GPs) are non-parametric models that can be employed to define prior distributions over functions~\cite{o1978curve,rasmussen2006gaussian}. Unlike parametric models, such as neural networks, GPs automatically adapt in complexity as more data are observed. 
They require only a small number of hyperparameters, making them suitable for training on small datasets, and they naturally quantify uncertainty in their predictions. 

To define a prior over a function $f: \mathcal{X} \rightarrow \mathbb{R}$, a GP assumes that, for every $n$ points ${\bf x}_1,\dots,{\bf x}_n \in  \mathcal{X}$, the probability distribution $p(f({\bf x}_1),\dots, f({\bf x}_n))$ is a multivariate Gaussian, with  mean vector $[\mu({\bf x}_1),\dots, \mu({\bf x}_n)]$ and  covariance matrix   $Cov(f({\bf x}_i),f({\bf x}_j)) = K({\bf x}_i,{\bf x}_j)$, for $i,j =1, \ldots, n$.
The functions $\mu({\bf x})$ and $K({\bf x},{\bf x}')$ are the mean  and the  kernel function of the GP, respectively. The mean function $\mu_{\boldsymbol{\theta}}({\bf x})$ and kernel $K_{\boldsymbol{\theta}}({\bf x},{\bf x}')$ usually depend on hyperparameters  $\boldsymbol{\theta} \in \Theta$. A typical example of kernel function  is the Matern $3/2$ kernel on $\mathbb{R}^d$. For ${\bf x},{\bf x}' \in  \mathcal{X} =\mathbb{R}^d$ it is defined as
\begin{equation}
\label{eq:Matern32}
K_{\boldsymbol{\theta}}({\bf x},{\bf x}')= \sigma^2\left(1+{\sqrt {3}}\gamma\right)\exp \left(-{{\sqrt {3}}\gamma}\right),
\end{equation}
where $\gamma=\sqrt{\sum_{i=1}^d (x_i-x'_i)^2/\ell_i^2}$ and $\boldsymbol{\theta}=[\ell_1,\dots,\ell_d,\sigma^2]$ includes the lengthscale hyperparameters $\ell_i >0$ (one for each dimension) and the scale parameter $\sigma^2$.
GPs, as priors over functions, have a natural Bayesian interpretation that makes them ideal for regression problems. If we assume that observations  are the sum of a true function evaluated at some inputs and Gaussian noise, i.e. $y_i = f({\bf x}_i) + \varepsilon_i$ with $\varepsilon_i\sim N(0, \varsigma^2)$ for $i=1, \ldots,n$, then  the posterior distribution of $f$ can be computed analytically. We can write the measurement model  compactly as the likelihood:
$$p(y_1, \ldots, y_n | f({\bf x}_1), \ldots, f({\bf x}_n))=N(\mathbf{y}_n| { f}(X), \varsigma^2I_n),$$
where $\mathbf{y}_n=[y_1,\dots,y_n]^\top$,  $X=[{\bf x}_1,\dots,{\bf x}_n]^\top$ and $I_n$ is the identity matrix of dimension $n$. Then, it can be easily shown  \cite{rasmussen2006gaussian} that the predictive posterior for the value of $f({\bf x}^*)$ at a new test point ${\bf x}^*\in \mathcal{X}$ is $GP(\mu_p,k_p)$, with mean and  kernel function:
\begin{align}
\label{eq:GPmean}
    \mu_p({\bf x}^*) &= K_{\boldsymbol{\theta}}({\bf x}^*,X) (K_{\boldsymbol{\theta}}(X,X)+\varsigma^2I_n)^{-1}\mathbf{y}_n \\
\nonumber
    k_p({\bf x}^*, {\bf x}^*) &= K_{\boldsymbol{\theta}}({\bf x}^*,{\bf x}^*)\\
    \label{eq:GPcov}
    &- K_{\boldsymbol{\theta}}({\bf x}^*,X)  (K_{\boldsymbol{\theta}}(X,X)+\varsigma^2I_n)^{-1}K_{\boldsymbol{\theta}}(X,{\bf x}^*),
\end{align}
where $K_{\boldsymbol{\theta}}(X,X)$ is a matrix whose ij-th element is defined as $(K_{\boldsymbol{\theta}}(X,X))_{ij}=K_{\boldsymbol{\theta}}({\bf x}_i,{\bf x}_j)$ (similar for $K_{\boldsymbol{\theta}}({\bf x}^*,X)$). The variance of the likelihood, $\varsigma^2$, is also considered to be a hyperparameter and included in $\boldsymbol{\theta}$. The values of the hyperparameters $\boldsymbol{\theta}$ are typically estimated by Maximum A-Posteriori (MAP) inference:
\begin{align}
\nonumber
&\arg\max_{\boldsymbol{\theta}}p( ({\bf x}_i,y_i)_{i=1}^n|\boldsymbol{\theta})p(\boldsymbol{\theta})\\
\label{eq:MAP}
&= \arg\max_{\boldsymbol{\theta}} N(\mathbf{y}_n,K_{\boldsymbol{\theta}}(X,X)+\varsigma^2 I_n)p(\boldsymbol{\theta}),
\end{align}
where $p(\boldsymbol{\theta})$ is a prior over the hyperparameters.

It is worth noticing that a GP, for  particular choices of the kernel function, can be viewed as a neural network with an infinite number of hidden units. This result was first shown  \cite{williams1996computing} for a single-layer and then generalised  \cite{NIPS2009_5751ec3e,Lee2018} for fully connected neural networks with $L$ hidden layers. In this case, the covariance function is defined as:
\begin{align}
\nonumber
k^{(0)}_{\boldsymbol{\theta}}({\bf x},{\bf x}')&= \sigma_b^2+\sigma_w^2 \frac{{\bf x}^\top {\bf x}'}{d}\\
\nonumber
k^{(L)}_{\boldsymbol{\theta}}({\bf x},{\bf x}')&= \sigma_b^2+\sigma_w^2 F_\phi\Big(k^{(L-1)}_{\boldsymbol{\theta}}({\bf x},{\bf x}'),\\
\label{eq:KNN}
&~~~~~~~~~~~~~~~k^{(L-1)}_{\boldsymbol{\theta}}({\bf x},{\bf x}),k^{(L-1)}_{\boldsymbol{\theta}}({\bf x}',{\bf x}')\Big),
\end{align}
where $ F_\phi$  is a deterministic function based on the activation function (e.g., ReLU). This kernel has only two  hyperparameters $\boldsymbol{\theta}=[\sigma_b^2,\sigma_w^2]$.

Taking inspiration from the multi-layer structure of deep neural networks, there is a line of work which instead considers stacking GPs, known as DeepGPs \cite{damianou2013deep,bui2016deep}, which can give rise to a richer class of probabilistic models beyond GPs.  
In contrast to highly parameterised deep models, DeepGPs learn a hierarchical representation non-parametrically, requiring only a few hyperparameters to optimise. However, since DeepGPs are not  GPs, their posterior cannot be computed exactly in regression, and various types of variational approximations are used to approximate it \cite{damianou2013deep,bui2016deep,salimbeni2017doubly}.

In problems with likelihoods different from the Gaussian, the posterior is not a GP. For  probit   and skew-normal   likelihoods (classification/preference learning problems), the posterior is a skew GP \cite{benavoli2020skew,benavoli2021preferential,benavoli2021}. For other types of likelihood, in general the posterior does not have a closed-form and  is approximated with a GP using three main approaches: (i) Laplace's approximation  \cite{mackay1996bayesian,williams1998bayesian}; (ii) Expectation Propagation \cite{minka2001family}; (iii) Kullback-Leibler divergence minimization \cite{opper2009variational}, including Variational approximation \cite{gibbs2000variational} as a particular case. 

A limitation of GPs is their high computational cost, with time complexity of $\mathcal{O}(n^3)$ and memory complexity of $\mathcal{O}(n^2)$, where $n$ represents the size of the training data. However, several established methods can scale GPs to $\mathcal{O}(n)$. These methods are usually known as \textit{sparse GP} methods and typically involve the use of inducing points \cite{quinonero2005unifying,snelson2006sparse}, in conjunction with variational inference \cite{pmlrv5titsias09a,Hensman2013,hernandez2016scalable}, or other well-known techniques \cite{bauer2016understanding,SCHURCH2020,schuch2023correlated}.

Since the inducing-point method is also pivotal for inference in DeepGPs, we briefly provide an overview of \textit{sparse GPs}. A {sparse GP} is built by augmenting the GP with a small number $m <n$ of inducing variables such that $\upsilon_i = f(\boldsymbol{\zeta}_i)$ with $\boldsymbol{\zeta}_i \in \mathcal{X},\upsilon_i \in \mathbb{R}$ for $i=1,\dots,m$ \cite{snelson2006sparse} and $f \sim GP(\mu({\bf x}),K_{\boldsymbol{\theta}}({\bf x},{\bf x}'))$. The pairs of inducing variables $\{\boldsymbol{\zeta}_i,\upsilon_i\}_{i=1}^m$ can be thought of as virtual observations. By conditioning the GP on the virtual observations $\{\boldsymbol{\zeta}_i,\upsilon_i\}_{i=1}^m$, we can compute the posterior predictive distribution $p(f(X)|Z,\Upsilon)$ at the real observations  $\{\mathbf{x}_i,y_i\}_{i=1}^n$, that is
\begin{equation}
\label{eq:GPsparse}
\begin{aligned}
p(f(X)|Z,\Upsilon)&=N(Q\Upsilon,K_{\boldsymbol{\theta}}(X,X)-QK_{\boldsymbol{\theta}}(Z,Z)Q^\top),\\
\Upsilon&\sim N({\bf 0},K_{\boldsymbol{\theta}}(Z,Z)),
\end{aligned}
\end{equation}
where $Q=K_{\boldsymbol{\theta}}(X,Z)K_{\boldsymbol{\theta}}^{-1}(Z,Z)$, $Z=[\boldsymbol{\zeta}_1,\dots,\boldsymbol{\zeta}_m]^\top$ and $\Upsilon=[\upsilon_1,\dots,\upsilon_m]^\top$. The inference problem
of sparse GP is to learn the hyperparameters ${\boldsymbol{\theta}}$ of the kernel and the virtual observations $\{\boldsymbol{\zeta}_i,\upsilon_i\}_{i=1}^m$ by maximising a lower bound
of the true log marginal likelihood  \cite{pmlrv5titsias09a}. The key
point of this approach is that the virtual observations are defined to be variational
parameters which are selected by minimizing
the Kullback-Leibler divergence between the
variational distribution and the exact posterior distribution over the latent function values. This prevents overfitting, since for $m \rightarrow n$ the variational approximation simply converges to the original GP posterior. Sparse GP have a time computational complexity of $\mathcal{O}(m^2n)$.

\section{Stochastic linear dynamical systems}
\label{sec:LTI}
A stochastic  linear time-invariant dynamical system is described by:
\begin{subequations}
\label{eq:linearSDE}
\begin{align}
\label{eq:linearSDE1}
d\z(t)&=A\z(t)dt+B\uu(t)dt+Ld\boldsymbol{\beta}(t),\\
\label{eq:linearSDE2}
y(t)&=C \z(t)+D\uu(t),
\end{align}
\end{subequations}
where $y(t) \in\mathbb{R}$ is the output  at time $t$, $\z(t) \in \mathbb{R}^{n_s}$ is the hidden state vector,  $\boldsymbol{\beta}(t) \in \mathbb{R}^{n_l}$ is Brownian motion with unit diffusion, $\uu(t) \in \mathbb{R}^{n_u}$ are $n_u$ deterministic input signals, and $A \in \mathbb{R}^{n_s \times n_s}$, $B  \in \mathbb{R}^{n_s \times n_u}$, $L  \in \mathbb{R}^{n_s \times n_l}$, $C  \in \mathbb{R}^{1 \times n_z}$ and $D  \in \mathbb{R}^{1 \times n_u}$ are time-invariant matrices. The initial state  
$\z(t_0) \sim N(\m(t_0),\Sigma(t_0, t_0))$
is assumed to be independent of the Brownian motion $\boldsymbol{\beta}(t)$. 

As shown for instance in~\cite{sarkka2019applied}, the solution to the linear stochastic differential equation~\eqref{eq:linearSDE1} is a  GP,  hereafter denoted by $\z(t) \sim GP(\m(t),\Sigma(t,t'))$. Furthermore, the mean $\m(t)$ and the covariance $\Sigma(t,t)$ of the state vector at time $t$ evolve according to the deterministic ordinary differential equations:
\begin{subequations}
\label{eq:mS}
\begin{align}
 \frac{d}{dt}  \m(t)&= A\m(t)+B\uu(t), \label{eq:mS1} \\
  \frac{d}{dt}  \Sigma(t,t)&= A\Sigma(t,t)+\Sigma(t,t)A^\top+ \label{eq:mS2}
  L L^\top,
\end{align}
\end{subequations}
with initial condition $\m(t_0), \Sigma(t_0,t_0)$ corresponding to the mean and covariance of $\z(t_0)$, respectively. The solutions to the differential equations above are:
\begin{subequations}    
\begin{align}
 \m(t)&= \Psi(t,t_0) \m(t_0)+  \int_{t_0}^t \Psi(t,\tau)B\uu(\tau)d\tau, \\
 \nonumber
  \Sigma(t,t)&= \Psi(t,t_0)\Sigma(t_0,t_0)\Psi(t,t_0)^\top\\
  &+\int_{t_0}^t \Psi(t,\tau)
  L L^\top \Psi(t,\tau)^\top d\tau,
\end{align}
\end{subequations}
where $\Psi(t,t')$ is the transition matrix:
\begin{equation}
\Psi(t,t') =e^{A(t-t')},
\end{equation}
and $e$ denotes the  \textit{matrix exponential}.
Furthermore, the covariance $\Sigma(t, t')$ between state vectors at distinct time instants $t,t'$ is given by: 
\begin{align}
\nonumber
 \Sigma(t,t')&=E[(\z(t)-\m(t))(\z(t')-\m(t'))^\top]\\
 &=\left\{\begin{array}{ll}
 \Sigma(t,t)\Psi^\top(t',t) & \text{if } t<t',\\
 \Psi(t,t') \Sigma(t',t') & \text{if } t\geq t'.\\
 \end{array}\right.
\end{align}

Owing to linearity of~\eqref{eq:linearSDE2}, the output $y(t)$ is also a GP:
\begin{align}
y(t) \sim GP(m_{\boldsymbol{\theta}}(t),S_{\boldsymbol{\theta}}(t,t')).
\end{align}
The covariance $S_{\boldsymbol{\theta}}(t,t')$ is given by:
\begin{align}
\label{eq:GP_cov_nosteady}
\nonumber
 S_{\boldsymbol{\theta}}(t,t')&=E[(y(t)-C\m(t)-D{\bf u}(t))\\
 \nonumber
 &~~~~~~~~(y(t')-C\m(t')-D{\bf u}(t))^\top]\\
 &=C\Sigma(t,t')C^\top,
\end{align}
while the mean is $m_{\boldsymbol{\theta}}(t)=C\m(t)+D {\bf u}(t)$.\footnote{To simplify the notation, we introduced an abuse of notation by denoting the mean of $y(t)$ by $m_{\boldsymbol{\theta}}(t)$
 (not in bold).}
Note that, the hyperparameters vector $\boldsymbol{\theta}$ of the GP will include the coefficients of the matrices $A,B,L,C,D$ as well as the variance of the noise $\varsigma^2$. 
 These parameters are estimated via MAP as described in \eqref{eq:MAP}.~\\

Assuming the LTI system is stable,~\eqref{eq:mS2} will have a steady-state solution $\Sigma_{\infty}$, which is the solution of the Lyapunov equation:
\begin{equation}
A\Sigma_{\infty}+\Sigma_{\infty} A^\top+L L^\top=0.
\end{equation}
Then, the GP covariance~\eqref{eq:GP_cov_nosteady} can be simplified to: 
\begin{equation}
\label{eq:covLyap}
\begin{aligned}
 S_{\boldsymbol{\theta}}(t,t')&=\left\{\begin{array}{ll}
 C\Sigma_{\infty} \Psi^\top(t',t)C^\top & \text{if } t<t',\\
 C\Psi(t,t') \Sigma_{\infty} C^\top & \text{if } t\geq t'.\\
 \end{array}\right.
\end{aligned}
\end{equation}

Hereafter, as an example, we report a well-known  connection between LTI systems and 1-dimensional Matern 3/2 kernels \cite{hartikainen2010kalman}, whose covariance function $S_{\boldsymbol{\theta}}(t,t')$ is
given in \eqref{eq:Matern32} with $\gamma=|t-t'|/\ell$ and $\ell>0$ being the lengthscale hyperparameter.~\\

\begin{example}
Consider the LTI system in \eqref{eq:linearSDE} with
\begin{equation}
\begin{array}{ll}
A=\begin{bmatrix}
    0 & 1\\
    -\frac{3}{\ell^2} & -2 \frac{\sqrt{3}}{\ell}
\end{bmatrix}, & 
L=\sigma\begin{bmatrix}
    0 \\
    \sqrt{\tfrac{36}{\ell^3}}
\end{bmatrix},\vspace{2mm}\\
C=\begin{bmatrix}
    1 & 0\\
\end{bmatrix},
\end{array}
\end{equation}
with $\btheta=[\sigma^2,\ell]$, $\ell>0$ and $B=\mathbf{0}$. The steady-state variance is 
\begin{equation}
\Sigma_{\infty}=\sigma^2\begin{bmatrix}
 \sqrt{3} & 0 \\
 0 & \frac{\sqrt{27}}{l^2} \\
\end{bmatrix}, 
\end{equation}
and
\begin{equation}
\Psi(t,t')=e^{A(t-t')}=\begin{bmatrix}
 e^{-\frac{\delta}{\ell}} (\frac{\delta}{\ell}+1) & \frac{\delta}{\sqrt{3}} e^{-\frac{\delta}{\ell}} \\
 -\frac{\sqrt{3}\delta}{\ell^2} e^{-\frac{\delta}{\ell}} & e^{-\frac{\delta}{\ell}} (1-\frac{\delta}{\ell}) \\
\end{bmatrix},
\end{equation}
with $\delta=\sqrt{3}(t-t')$. 
 Therefore, it is easy to verify that 
\begin{equation}
\begin{aligned}
&C\Sigma_{\infty}(e^{A(t-t')})^\top C^\top=Ce^{A(t'-t)} \Sigma_{\infty}C^\top\\
&=\sqrt{3}\sigma^2 e^{-\tfrac{|\delta|}{\ell}} \left(\tfrac{|\delta|}{\ell}+1\right),
\end{aligned}
\end{equation}
which is proportional to  $K_{\boldsymbol{\theta}}(t,t')$ 
given in \eqref{eq:Matern32}.\\
\end{example}

\begin{remark}
Having illustrated the equivalence between stochastic LTI systems and GPs, $y(t) \sim GP(m(t),S(t,t'))$, given observations $y(t_1),\dots,y(t_n)$ at time $t_1,\dots,t_n$, the predictive posterior over $y$ can be computed as described in \eqref{eq:GPmean}--\eqref{eq:GPcov}. This allows us to calculate the predictive distribution $p(y(t^*) | data)$ for each time $t^*$. This computation is equivalent to performing filtering and smoothing using recursive methods such as the Kalman filter and smoother \cite{carron2016machine}. In GPs, however, this is achieved in a non-sequential manner without computing an estimate of the state, yet the resulting predictive distribution remains the same. Also, note that the estimate of the hyperparameters via MAP estimation is equivalent to what is commonly done in Kalman filtering. This is because the marginal likelihood in \eqref{eq:MAP} coincides with the marginal likelihood computed via the Kalman filter.
\end{remark}

State-space representations for other common kernels, in addition to the Matern kernel, are provided in \cite{hartikainen2010kalman,loper2021general}.
In this paper, we focus on LTI dynamical systems. However, the connection between stochastic linear dynamical system and GPs also extends to time-variant cases, as shown for instance in \cite{solin2014explicit,benavoli2016state}. For example, \cite{benavoli2016state} derived the linear time-variant representation of the single-layer neural network kernel defined in \cite{williams1996computing}.

It is important to note that a GP defined by a stochastic linear dynamical system is indexed by time, meaning its mean and covariance functions depend on a one-dimensional covariate, that is time. In contrast, in Section \ref{sec:GP}, we considered a GP indexed by a vector $\mathbf{x} \in \mathbb{R}^d$. Such GPs cannot be represented by stochastic differential equations. However, alternative representations exist, involving stochastic partial differential equations \cite{sarkka2019applied}. This distinction is important for the main objective of this paper which is building a hierarchical composition of GPs as we will explain in Section \ref{sec:dynoGP}.

\subsection{Efficient computation of the mean and covariance}
\label{sec:efficient}
Hereafter, we will discuss how  to efficiently compute the mean and covariance functions in 
$GP(m_{\boldsymbol{\theta}}(t),S_{\boldsymbol{\theta}}(t,t'))$.
We will assume that time has been discretised, $t=k\delta$, where $k=0,1,2,\dots,n$ and $\delta>0$ is the sampling time. 
Under this assumption, we can reduce the computation of the $n \times n$  elements of the covariance function $S(t,t')$ for $t,t' \in \{k\delta:~0,1,2,\dots,n-1\}$ to the computation of $n$ elements. Indeed, for any $t$, from \eqref{eq:covLyap} we have that 
\begin{equation}
\label{eq:covydelta}
S(t,t+k\delta)=C\Sigma_{\infty}(e^{Ak \delta })^\top C^\top= C\Sigma_{\infty}(\bar{A}^k)^\top C^\top,
\end{equation}
where $\bar{A}=e^{A \delta }$. Therefore, the $n \times n$ covariance matrix can be constructed by computing the $n$ elements $C\Sigma_{\infty}(\bar{A}^k)^\top C^\top$ for $k = 0, \dots, n-1$ and leveraging stationarity. This implies that $S(t, t')$ depends only on $|t - t'|$, and consequently, the matrix $S(T, T)$, with $T = [t_1, \dots, t_n]^\top$, contains at most $n$ distinct elements.

A similar approach can be applied to the computation of the mean.  By assuming a zero-order hold (ZOH) interpolation between discretised inputs, we have that 
\begin{equation}
\label{eq:meanydelta}
m(k\delta)=C\left(\bar{A}^{k}{\bf m}(0)+ \sum_{i=1}^{k} \bar{A}^{i-1} \bar{B}\uu(k\delta)\right)+ D \uu(k\delta), 
\end{equation}
where $\bar{B}=(\bar{A}-I)A^{-1}B$ is the discretised $B$ matrix. In the following, since we will only consider stable LTI systems, we will set $\m(0)=\mathbf{0}$.\\
In both \eqref{eq:covydelta} and \eqref{eq:meanydelta}, the bottleneck in the computation is determined by $e^{Ak \delta }=\bar{A}^k$.  A way to further accelerate the construction of the mean and kernel function is by restricting the structure of the $A$ matrix.

\subsubsection{Complex diagonal}
An efficient calculation of the mean and kernel function  can be obtained by working with a dynamical system with diagonal matrix $A$. In particular, we consider a diagonal matrix with complex conjugate eigenvalues. 
Consider the following system with $n_s=2$  states and complex-conjugate eigenvalues: 
\begin{equation}
\label{eq:complexsys}
\begin{aligned}
d {\bf z}_{c}(t) &= 
\overbrace{
 \begin{bmatrix}
\lambda   & 0 \\
0  &  \lambda^\dagger 
\end{bmatrix}
}^{A_{c}}
{\bf z}_{c}(t)dt + 
\overbrace{
\begin{bmatrix}
B\\
B^\dagger
\end{bmatrix}
}^{B_{c}}
{\bf u}(t)dt+\overbrace{
\begin{bmatrix}
L\\
L^\dagger
\end{bmatrix}
}^{L_{c}}
d\boldsymbol{\beta}(t), \\
y_c(t) & = 
\overbrace{\begin{bmatrix}c  & c^\dagger\end{bmatrix}}^{C_{c}} {\bf z}_{c}(t)+D{\bf u}(t),
\end{aligned}
\end{equation}

with $\lambda,c \in \mathbb{C}$, $B  \in \mathbb{C}^{2 \times n_u}$, $L  \in \mathbb{C}^{2 \times n_l}$, ${\bf z}_{c}(0)=[\chi,\chi^\dagger]^\top$ with $\chi \in \mathbb{C}$ and ${\bf u}(t),\boldsymbol{\beta}(t)$ are real-valued. We further assume that the real part of $\lambda$ is less than zero for stability.

We can transform the system \eqref{eq:complexsys}  
into the real form:
\begin{equation}
\label{eq:complexsysreal}
\begin{aligned}
d {\bf z}_{r}(t) &= 
\overbrace{
 \begin{bmatrix}
\Re[\lambda]   &\Im[\lambda]  \\
-\Im[\lambda]  & \Re[\lambda] 
\end{bmatrix}
}^{A_{r}}
{\bf z}_{r}(t)dt + 
\overbrace{
\begin{bmatrix}
\sqrt{2} \Re [B] \\
-\sqrt{2} \Im[B]
\end{bmatrix}
}^{B_{r}}
{\bf u}(t)dt\\
&+\overbrace{
\begin{bmatrix}
\sqrt{2} \Re [L] \\
-\sqrt{2} \Im[L]
\end{bmatrix}
}^{L_{r}}
d\boldsymbol{\beta}(t) \\
y_r(t) & = 
\overbrace{\begin{bmatrix}\sqrt{2} \Re[c] & - \sqrt{2} \Im[c]\end{bmatrix}}^{C_{r}} {\bf z}_{r}(t)+D{\bf u}(t),
\end{aligned}
\end{equation}
where ${\bf z}_{r}(t) \in \mathbb{R}^2$, with the transformation 
\begin{align}
{\bf z}_r &= J {\bf z}_{c} \\
J &= \tfrac{1}{\sqrt{2}} \begin{bmatrix}
1 & 1 \\
\iota & -\iota
\end{bmatrix}, ~~J^{-1} = J^H= 
\tfrac{1}{\sqrt{2}} 
\begin{bmatrix}
1 & -\iota \\
1 & \iota
\end{bmatrix},
\end{align}
with $\iota$ being the complex unit  and the superscript $^H$  denotes the Hermitian transpose. Note that, 
$\Re[x]$ and $\Im[x]$ denote the real part and, respectively, imaginary part of $x$.
By exploiting $A_c=J^HA_rJ$ and $L_c=J^H L_r$, it is easy to prove that the solutions of the Lyapunov equations for the two systems:
\begin{align}
\label{eq:lyapjordan}
A_{c} S^{(c)}_{\infty} + S^{(c)}_{\infty}  A_{c}^H + L_{c}L^H_{c} &= 0,\\
A_{r} S^{(r)}_{\infty}  + S^{(r)}_{\infty}  A_{r}^H+ L_{r}L^H_{r} &= 0,
\end{align}
 are related by $S^{(r)}_{\infty} = J S^{(c)}_{\infty} J^H$. Therefore, we have that {\small\begin{equation}
\label{eq:covcompelx}
\begin{aligned}
 &S_{\boldsymbol{\theta}}(t,t')=\\
 &\left\{\begin{array}{ll}
 C_{r} S^{(r)}_{\infty} \Psi_r^
 H(t',t) (C^{(r)})^H=C_{c} S^{(c)}_{\infty} \Psi_c^
 H(t',t) (C^{(c)})^H& \text{if } t<t'\\
 C_{r} \Psi_r(t,t') S^{(r)}_{\infty} (C^{(r)})^H=C_{c} \Psi_c(t,t')S^{(c)}_{\infty} (C^{(c)})^H& \text{if } t\geq t'\\
 \end{array}\right.
\end{aligned}
\end{equation}}

where
\begin{align}
\Psi_c(t,t')&=e^{A_c(t-t')},\\
\Psi_r(t,t')&=e^{A_r(t-t')}.
\end{align}
Note that, since $A_c$ is diagonal and from the properties of the matrix exponential, we have that
$\Psi_r(t,t')=J\Psi_c(t,t')J^H$, which allows one to easily verify \eqref{eq:covcompelx}.
This shows that, for the system \eqref{eq:complexsysreal}, we can efficiently compute the elements of the kernel $S_{\boldsymbol{\theta}}(t, t')$ by working in the complex domain, where the corresponding diagonal dynamic matrix $A_c$ enables efficient computation of the matrix exponential. Indeed, it is straightforward to derive the following results.\\

\begin{lemma}
\label{lem:1}
The solution of the Lyapunov equation \eqref{eq:lyapjordan} is:
\begin{equation}
\Sigma_{\infty}^{(c)}=-\begin{bmatrix}
\frac{LL^H}{\lambda+\lambda^\dagger}  & \frac{LL^\top}{2\lambda} \\
\frac{L^\dagger L^H}{2\lambda^\dagger}  & \frac{L^\dagger L^\top}{\lambda+\lambda^\dagger} \\
\end{bmatrix}.
\end{equation}
\end{lemma}
The proof of this lemma and next propositions can be found in Appendix \ref{app:proofs}. Hence, the kernel function has the following analytical form.\\

\begin{proposition}
\label{prop:1}
The kernel \eqref{eq:covcompelx} is equal to:
\begin{equation}
S_{\boldsymbol{\theta}}(t,t')=\left\{\begin{array}{ll}\begin{aligned}
&-e^{\lambda^\dagger \tau}\left(\tfrac{cLL^Hc^\dagger}{\lambda+\lambda^\dagger}  +\tfrac{c^\dagger L^\dagger L^Hc^\dagger }{2\lambda^\dagger} \right)\\
&~-e^{\lambda \tau}\left(\tfrac{cLL^\top c}{2\lambda}  +\tfrac{c^\dagger L^\dagger L^\top c }{\lambda+\lambda^\dagger} \right)
\end{aligned} & \text{if } t'>t,\\
& \\
\begin{aligned}
&-e^{\lambda^\dagger \tau}\tfrac{cLL^Hc^\dagger}{\lambda+\lambda^\dagger}  -e^{\lambda \tau}\tfrac{c^\dagger L^\dagger L^Hc^\dagger }{2\lambda^\dagger} \\
&-e^{\lambda^\dagger  \tau}\tfrac{cLL^\top c}{2\lambda}  -e^{\lambda \tau}\tfrac{c^\dagger L^\dagger L^\top c }{\lambda+\lambda^\dagger} 
\end{aligned}& \text{if } t'<t,
\end{array}\right.
\end{equation}
with $\tau=|t'-t|$.
\end{proposition}

For the computation of the mean function $m_{\boldsymbol{\theta}}(t)$, we can  show that the following one-dimensional deterministic dynamical system has the same mean function as the one in \eqref{eq:complexsys}.~\\

\begin{proposition}
\label{prop:2}
The systems in  \eqref{eq:complexsys} and \eqref{eq:complexsys1} have the same mean function for $y(t)$.
\begin{equation}
\label{eq:complexsys1}
\begin{aligned}
d x_{c}(t) &= 
\overbrace{
 \begin{bmatrix}
\lambda   \\ 
\end{bmatrix}
}^{A_{c}}
x_{c}(t)dt + 
\overbrace{
\begin{bmatrix}
B
\end{bmatrix}
}^{B_{c}}
{\bf u}(t)dt \\
y(t) & = 2\Re\left(
\overbrace{\begin{bmatrix}c  \end{bmatrix}}^{C_{c}} x_{c}(t)\right)+D{\bf u}(t).
\end{aligned}
\end{equation}
The mean function is:
\begin{equation}
\begin{aligned}
&{m}_{\boldsymbol{\theta}}(k\delta)\\
&=2\Re\left(ce^{\lambda k\delta}{\tilde m}(0)+ \sum_{i=1}^{k} \tfrac{c}{\lambda}e^{\lambda (i-1)\delta} (e^{\lambda \delta}-1)B\uu(i\delta)\right)\\
&+ D \uu(k\delta),
\end{aligned}
\end{equation}
where ${\tilde m}(0)$ denotes the first component of ${\bf m}(0)$.
\end{proposition}

In \eqref{eq:complexsys}, we defined a dynamical system with a two-dimensional state. We can  build a $n_s$-dimensional system using an additive composition of $n_s/2$ two-dimensional systems. For instance, for $n_s=4$, this results into the following 4-dimensional system:
\begin{equation}
\label{eq:complexsys2s}
\scalebox{0.85}{$
\begin{aligned}
d {\bf z}_{c}(t) &= 
\overbrace{
 \begin{bmatrix}
\lambda_1   & 0 & 0 & 0 \\
0  &  \lambda_1^\dagger  & 0 & 0\\
0  &  0 & \lambda_2  & 0\\
0  &  0  & 0 & \lambda_2^\dagger \\
\end{bmatrix}
}^{A_{c}}
{\bf z}_{c}(t)dt + 
\overbrace{
\begin{bmatrix}
B_1\\
B_1^\dagger\\
B_2\\
B_2^\dagger
\end{bmatrix}
}^{B_{c}}
{\bf u}(t)dt+\overbrace{
\begin{bmatrix}
L_1\\
L_1^\dagger\\
L_2\\
L_2^\dagger
\end{bmatrix}
}^{L_{c}}
d\boldsymbol{\beta}(t), \\
y(t) & = 
\overbrace{\begin{bmatrix}c_1  & c_1^\dagger & c_2 & c_2^\dagger\end{bmatrix}}^{C_{c}} {\bf z}_{c}(t)+D{\bf u}(t).
\end{aligned}$}
\end{equation}

\begin{proposition}
\label{prop:3}
Consider   the additive composition of   $n_s/2$ two-dimensional systems of type  \eqref{eq:complexsys}, then the output $y(t)$ is distributed as $GP(m_{\boldsymbol{\theta}}(t),S_{\boldsymbol{\theta}}(t,t'))$, with mean and covariance function:
\begin{equation}
\label{eq:summean}
\begin{aligned}
{m}_{\boldsymbol{\theta}}(k\delta)&=\sum_{j=1}^{n_s} 2\Re\Bigg(c_je^{\lambda_j k\delta}{\tilde m}_j(0)\\
&~+ \sum_{i=1}^{k} \frac{c_je^{\lambda_j (i-1)\delta}(e^{\lambda_j \delta}-1)}{\lambda_j}B_j\uu(i\delta)\Bigg)+ D \uu(k\delta).
\end{aligned}
\end{equation}
and
\begin{equation}
\label{eq:sumcov}
\scalebox{0.9}{$
S_{\boldsymbol{\theta}}(t,t')=\left\{\begin{array}{ll}\begin{aligned}
&\sum_{i=1}^{n_s} \Bigg(-e^{\lambda_i^\dagger \tau}\left(\tfrac{c_iL_iL_i^Hc_i^\dagger}{\lambda_i+\lambda_i^\dagger}  +\tfrac{c_i^\dagger L_i^\dagger L_i^Hc_i^\dagger }{2\lambda_i^\dagger} \right)\\
&~-e^{\lambda_i \tau}\left(\tfrac{c_iL_iL_i^\top c_i}{2\lambda_i}  +\tfrac{c_i^\dagger L_i^\dagger L_i^\top c_i }{\lambda_i+\lambda_i^\dagger} \right)\Bigg)
\end{aligned} & \text{if } t'>t,\\
& \\
\begin{aligned}
&\sum_{i=1}^{n_s} \Bigg(-e^{\lambda^\dagger \tau}\tfrac{cLL^Hc^\dagger}{\lambda+\lambda^\dagger}  -e^{\lambda \tau}\tfrac{c^\dagger L^\dagger L^Hc^\dagger }{2\lambda^\dagger} \\
&-e^{\lambda^\dagger  \tau}\tfrac{cLL^\top c}{2\lambda}  -e^{\lambda \tau}\tfrac{c^\dagger L^\dagger L^\top c }{\lambda+\lambda^\dagger} \Bigg)
\end{aligned}& \text{if } t'<t,
\end{array}\right.$}
\end{equation}
with $\tau=|t'-t|$. 

\end{proposition}

We will now present an illustrative example demonstrating the use of a GP with the above derived mean and covariance function for system identification.\\

\begin{example}
\label{ex:2}
We considered a two-dimensional input $\mathbf{u}(t) = [\sin(3\pi t), \sin(5\pi t)]$ and simulated a signal $y(t)$ using an LTI system similar to \eqref{eq:complexsys2s}, but with a 10-dimensional state (the additive composition of $5$ two-dimensional systems of type  \eqref{eq:complexsys}). The parameters $\lambda_i$, $L_i$, $c_i \in \mathbb{C}$, and $B_i \in \mathbb{C}^2$ for $i=1,\dots,5$ were randomly generated. The standard deviation $\varsigma$ of the measurement noise was set to 10\% of the standard deviation of the LTI system's output. A sampling time of $\delta = 0.01$ was used, and data were generated over the interval $t = 0$ to $t = 25$. For system identification, we used the data for $t \leq 15$, shown in Figure \ref{fig:ex1}-top. The remaining data for $t > 15$ were generated without process noise and used for testing the model. We assumed $y \sim GP(m_{\boldsymbol{\theta}}(\mathbf{u}(t)), S_{\boldsymbol{\theta}}(t, t'))$, where the mean and covariance functions are defined in \eqref{eq:summean}--\eqref{eq:sumcov}, and $n_s = 20$. The vector $\boldsymbol{\theta}$ includes all the model's hyperparameters, namely: $\lambda_i$, $L_i$, $c_i \in \mathbb{C}$, and $B_i \in \mathbb{C}^2$ for $i = 1, \dots, 10$, and $\varsigma$. For generality, we assumed a system dimension of 20, although the true system used to generate the data had a dimension of 10.\\
We estimated the hyperparameters $\boldsymbol{\theta}$ using MAP, as described in \eqref{eq:MAP}, and computed the predictive posterior distribution, which is a GP. The posterior mean and covariance functions were calculated as in \eqref{eq:GPmean}–\eqref{eq:GPcov}. Figure \ref{fig:ex1}-bottom displays the predictive posterior and the 95\% credible region for $5 \leq t \leq 25$. The figure includes predictions for the training data interval $5 \leq t \leq 15$ and predictions for the unseen data interval $15 \leq t \leq 25$.
The model maintains good accuracy, as evidenced by its comparison with the true test data. At the beginning of the prediction task ($t = 5$), the prediction error is relatively large due to the transient phase. Since the GP model does not directly estimate the state, predictions assume an initial state of zero, requiring a few time steps for the model to converge.
   \begin{figure*}
    \centering
        \includegraphics[width=16cm]{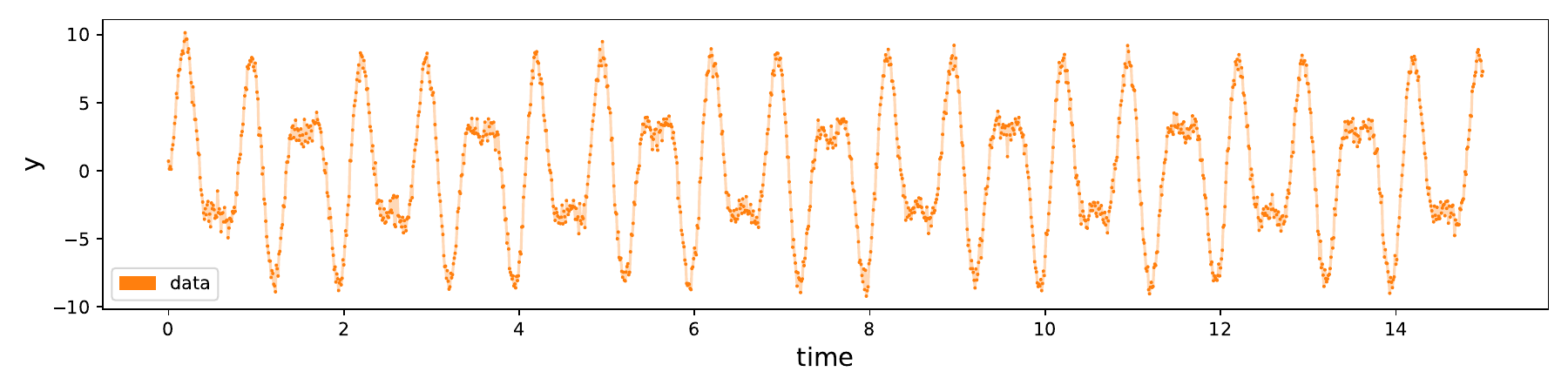}\vspace{-0.2cm}\\
\includegraphics[width=16cm]{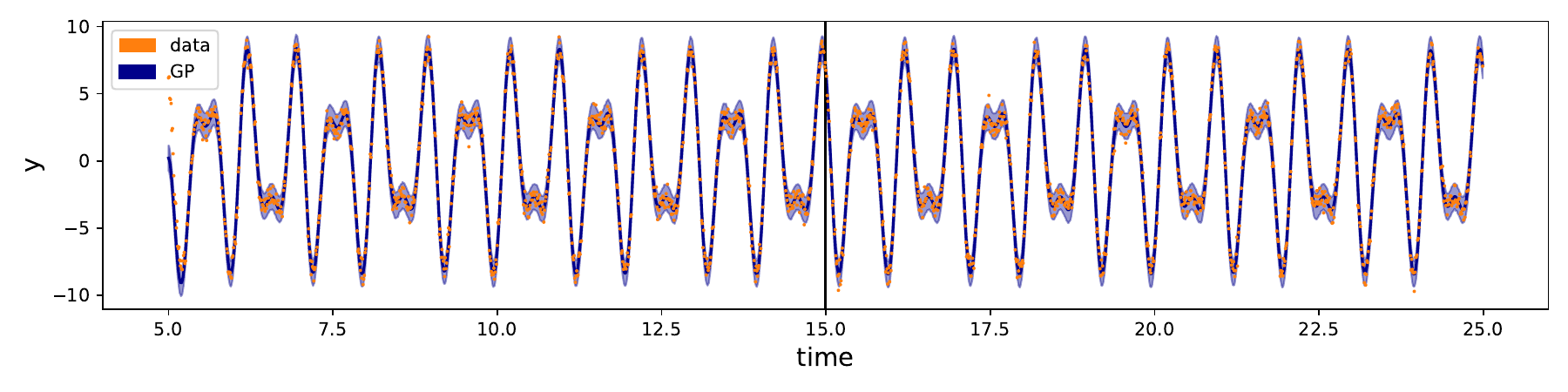}
\caption{Example: system identification of an LTI system using Gaussian Processes. The top plot shows the training data, while the bottom plot reports the predictions along with the 95\% credible interval in blue. The values to the left of the vertical bar represent the last 1000 values of the training set, while those to the right correspond to the test set.}
\label{fig:ex1}
    \end{figure*}
\end{example}

~\\
\begin{remark}
LTI deterministic state-space models, such as HIPPO \cite{gu2020hippo,gupta2022diagonal,gu2022parameterization}, have recently gained popularity in deep learning as alternatives to self-attention in transformers. These works primarily focus on parametrising the state-transition matrix with structures that enable fast computation of the mean of the deterministic LTI system. This focus has led to the adoption of diagonal state transition matrices or diagonal plus low-rank structures. For the dynamic layer in dynoGP (discussed in the next Section), we have exclusively considered diagonal matrices. These not only yield expressive state-space models but also facilitate efficient computation of both the mean and the covariance function of the GP, the latter requiring the solution of the Lyapunov equation. As we have shown these quantities can be computed in closed form in the diagonal case.
\end{remark}

\section{Deep dynamical GPs}
\label{sec:dynoGP}
In this section, we introduce deep dynamical GPs (dynoGPs). First, we present the following notation. We refer to the GPs introduced in Section \ref{sec:GP} -- where the covariate is a multidimensional vector ${\bf x} \in \mathbb{R}^d$ that does not include time -- as  \textit{static GPs}, denoted as $GP(\mu_{\boldsymbol{\theta}}, K_{\boldsymbol{\theta}})$. We instead refer to the GPs introduced in Section \ref{sec:LTI}, where the covariate is time, as  \textit{dynamic GPs}, denoted as $GP(m_{\boldsymbol{\theta}}, S_{\boldsymbol{\theta}})$, which can equivalently be rewritten as stochastic linear dynamical systems.

A dynoGP is constructed by stacking layers of static and dynamic GPs in cascade.
As an example, we will  focus on the Wiener model, which is defined as the cascade connection of an LTI dynamical model $g_t$ followed by a static nonlinearity $f$. In other words, a MISO (Multiple Input, Single Output) Wiener model is described by the following equation
\begin{equation}
\label{eq:Wiener}
y(t) = f \big(g_t\big({\bf u}(t)\big)\big) + \epsilon_t,
\end{equation}
where $\epsilon_t \sim N(0,\varsigma^2)$ is the measurement noise. A dynoGP model for \eqref{eq:Wiener} assumes that 
\begin{align}
\label{eq:gwiener}
g_t &\sim GP(m_{\boldsymbol{\theta}}({\bf u}(t)),S_{\boldsymbol{\theta}}(t,t')),\\
\label{eq:fwiener}
f &\sim GP(\mu_{\boldsymbol{\theta}}(g_t),K_{\boldsymbol{\theta}}(g_t,g'_t)),
\end{align}
where $\boldsymbol{\theta}$ is the vector of all hyperparameters  including $\varsigma^2$.
The GP in \eqref{eq:gwiener} describes the distribution of trajectories $g_t$ of an LTI dynamical system with mean function $m_{\boldsymbol{\theta}}({\bf u}(t))$ and covariance function $S_{\boldsymbol{\theta}}(t,t')$. For each generated trajectory $g_t$, $y(t)$ is assumed to be obtained by first sampling a static nonlinear function $f(g_t)$ from the GP in \eqref{eq:fwiener} and then adding i.i.d.\ Gaussian noise $\epsilon_t$.

We introduce the following pictorial diagrams to define the architecture of a dynoGP. The diagram below represents a Wiener architecture. The first block corresponds to a dynamic GP, while the second block represents a static GP. The signature $n_s, n_b, n_l, n_d$ in the dynamic block specifies the number of columns for the matrices $A, B, L, D$, respectively. A value of zero for any of these indicates that the corresponding matrix is zero.

{\centering
\includegraphics[width=9cm,trim={1cm 0 0 0},clip]{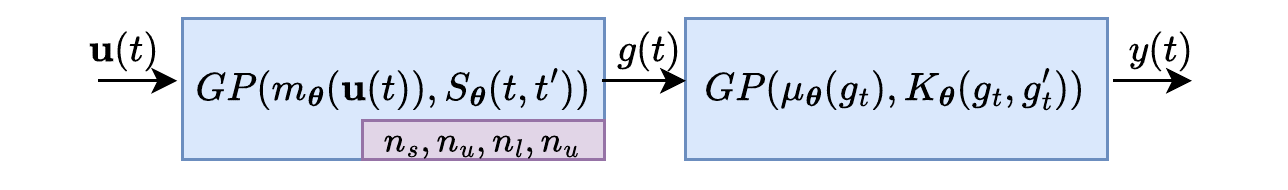}}

Given a set of (i) observations $y(t)$; (ii) values of the inputs ${\bf u}(t)$; at time $t=k\delta$ for $k=1,\dots,n$, the problem of inference in dynoGP consists of computing the posterior $p(f(g_t)|\{y(i\delta),{\bf u}(i\delta)\}_{i=1}^n)$.
Given the model \eqref{eq:Wiener}--\eqref{eq:fwiener},  we first start defining the joint distribution:

\begin{equation}
\label{eq:joint}
\scalebox{0.9}{$
\begin{aligned}
&p\Big(y(\delta),\dots,y(n\delta),f \big(g_{\delta}\big({\bf u}(\delta)\big)\big),\dots,f \big(g_{n\delta}\big({\bf u}(n\delta)\big),\\
&~~~g_{\delta}\big({\bf u}(\delta)\big),\dots,g_{n\delta}\big({\bf u}(n\delta)\big)\Big|{\bf u}(\delta),\dots,{\bf u}(n\delta)\big)\Big)\\
&=p(y(\delta),\dots,y(n\delta)|f \big(g_{\delta}\big({\bf u}(\delta)\big)\big),\dots,f \big(g_{n\delta}\big({\bf u}(n\delta)\big)\big))\\
&p(f \big(g_{\delta}\big({\bf u}(\delta)\big)\big),\dots,f \big(g_{n\delta}\big({\bf u}(n\delta)\big)|g_\delta\big({\bf u}(\delta)\big),\dots,g_{n\delta}\big({\bf u}(n\delta)\big))\\
&p(g_{\delta}\big({\bf u}(\delta)\big),\dots,g_{n\delta}\big({\bf u}(n\delta)\big)|{\bf u}(\delta),\dots,{\bf u}(n\delta)),\\
\end{aligned}$}
\end{equation}

which captures the generative process of the model, describing how the input ${\bf u}$ produce $y$.

The first term in \eqref{eq:joint} is the likelihood:
\begin{equation}
\label{eq:likedeep}
\begin{aligned}
&p(y(\delta),\dots,y(n\delta)|f \big(g_\delta\big({\bf u}(\delta)\big)\big),\dots,f \big(g_{n\delta}\big({\bf u}(n\delta)\big)\big))\\
&=N\left(\begin{bmatrix}
    y(\delta)\\\vdots\\y(n\delta)
\end{bmatrix};\begin{bmatrix}
     f\big(g_\delta\big({\bf u}(\delta)\big)\big)\\
     \vdots\\
     f \big(g_{n\delta}\big({\bf u}(n\delta)\big)\big)
\end{bmatrix}, \varsigma^2 I_n\right),\\
\end{aligned}
\end{equation}
with $I_n$ being the identity matrix of dimension $n$. The second term is the static GP (outer layer) and, therefore, the conditional probability is a multivariate normal distribution:
\begin{equation}
\label{eq:outerdeep}
\begin{aligned}
&\scalebox{0.9}{$p(f \big(g_{\delta}\big({\bf u}(\delta)\big)\big),\dots,f \big(g_{n\delta}\big({\bf u}(n\delta)\big)|g_\delta\big({\bf u}(\delta)\big),\dots,g_{n\delta}\big({\bf u}(n\delta)\big))$}\\
&=\scalebox{0.9}{$N\left(\begin{bmatrix}
     f\big(g_{\delta}\big({\bf u}(\delta)\big)\big)\\
     \vdots\\
     f \big(g_{n\delta}\big({\bf u}(n\delta)\big)\big)
\end{bmatrix};\begin{bmatrix}
     \mu_{\boldsymbol{\theta}}\big(g_\delta\big({\bf u}(\delta)\big)\big)\\
     \vdots\\
    \mu_{\boldsymbol{\theta}}\big(g_{n\delta}\big({\bf u}(n\delta)\big)\big)
\end{bmatrix}, K({\bf g},{\bf g})\right),$}\\
\end{aligned}
\end{equation}
where ${\bf g}=[g_\delta\big({\bf u}(\delta)\big),\dots,g_{n\delta}\big({\bf u}(n\delta)\big)]^\top$. Finally, the last term is the dynamic  GP (inner layer), leading to the multivariate normal distribution:
\begin{equation}
\label{eq:outerdeep1}
\begin{aligned}
&p(g_\delta\big({\bf u}(\delta)\big),\dots,g_{n\delta}\big({\bf u}(n\delta)\big)|{\bf u}(\delta),\dots,{\bf u}(n\delta))\\
&=N\left(\begin{bmatrix}
     g_\delta\big({\bf u}(\delta)\big)\\
     \vdots\\
    g_{n\delta}\big({\bf u}(n\delta)\big)
\end{bmatrix};\begin{bmatrix}
     m_{\boldsymbol{\theta}}\big({\bf u}(\delta)\big)\\
     \vdots\\
    m_{\boldsymbol{\theta}}\big({\bf u}(n\delta)\big)
\end{bmatrix}, S_{\boldsymbol{\theta}}(T,T)\right).\\
\end{aligned}
\end{equation}
Our goal is to compute the posterior distribution $p(f \big(g_{\delta}\big({\bf u}(\delta)\big)\big),\dots,f \big(g_{n\delta}\big({\bf u}(n\delta)\big)|\{y(i\delta),{\bf u}(i\delta)\}_{i=1}^n\big)$ and the posterior predictive process. Since no analytical solution exist for both the posteriors we compute a stochastic variational approximation as in \cite{salimbeni2017doubly}. 

First, we introduce inducing points (virtual observations) for each GP, as discussed in Section \ref{sec:GP}. These inducing points enable us to compute the posterior predictive distribution for each GP, which remains a GP when conditioned on the inducing points. At this stage, we can compute the predictive posterior for $f(g_t)$ by simply sampling from these GPs. As a result, the posterior at the last layer is approximated by a Gaussian mixture. The inducing points and hyperparameters $\boldsymbol{\theta}$ are optimized using variational inference, achieved by maximizing an evidence lower bound.  To deal with large datasets, we further subsample the data in minibatches. This  allows the model to scale effectively to arbitrarily large datasets. Additional details about the variational inference procedure are provided in Appendix \ref{app:svi}.\\

\begin{example}
\label{ex:3}
For illustration, we again considered the LTI system in Example \ref{ex:2}. However, in this case, we introduced a nonlinearity to the output before measuring it. Specifically, we generated the observations as $y'(t) = (y(t))^+ + \epsilon_t$, where $(\cdot)^+$ represents the positive part. The training data are shown in Figure \ref{ex:2}-top, where the effect of nonlinearity can be easily seen.
We then use a dynoGP model to identify this system. We first considered the same model as in Example \ref{ex:2}, a dynamic GP (that is, an LTI system). Figure \ref{ex:2}-center shows the predictive posterior mean and 95\% credible region for the last part of the training data and for the test data ($t>15$). It can be observed that the model struggles to handle the nonlinearity, resulting in larger uncertainty for both the training and test data, and clearly incorrect predictions for the test data. Figure \ref{ex:2}-bottom shows a dynoGP with a Wiener architecture, which incorporates the same dynamic GP as before, followed by a static GP. This approach significantly improves the model's accuracy as expected.\\

    \begin{figure*}
    \centering
        \includegraphics[width=16cm]{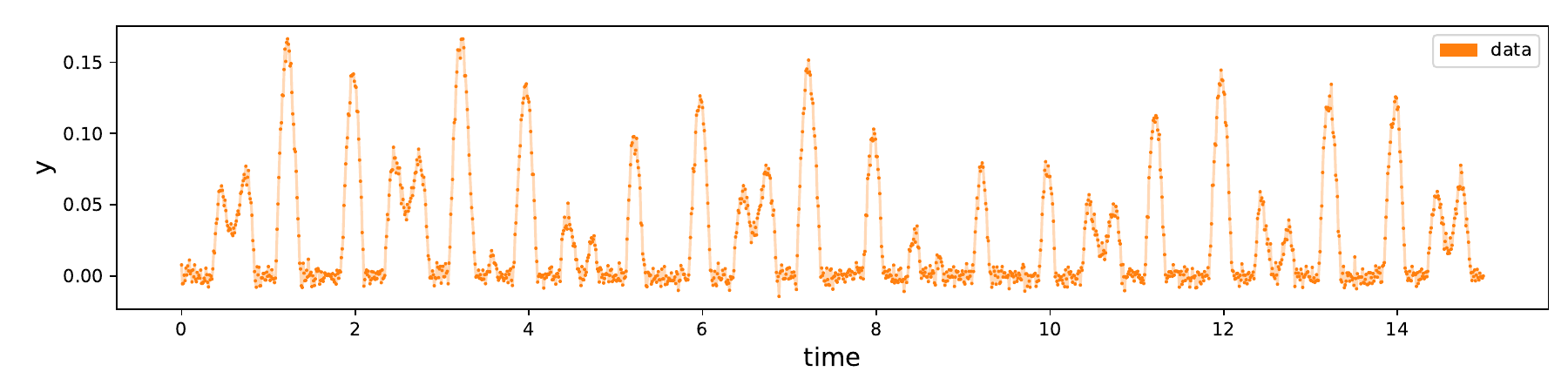}\vspace{-0.1cm}
\includegraphics[width=16cm]{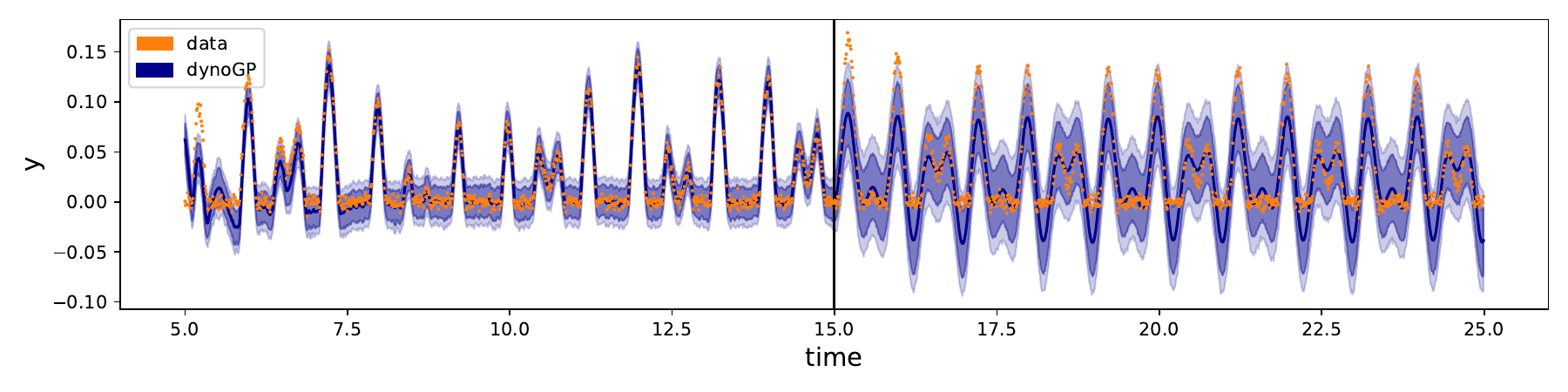}
\vspace{-0.1cm}
\includegraphics[width=16cm]{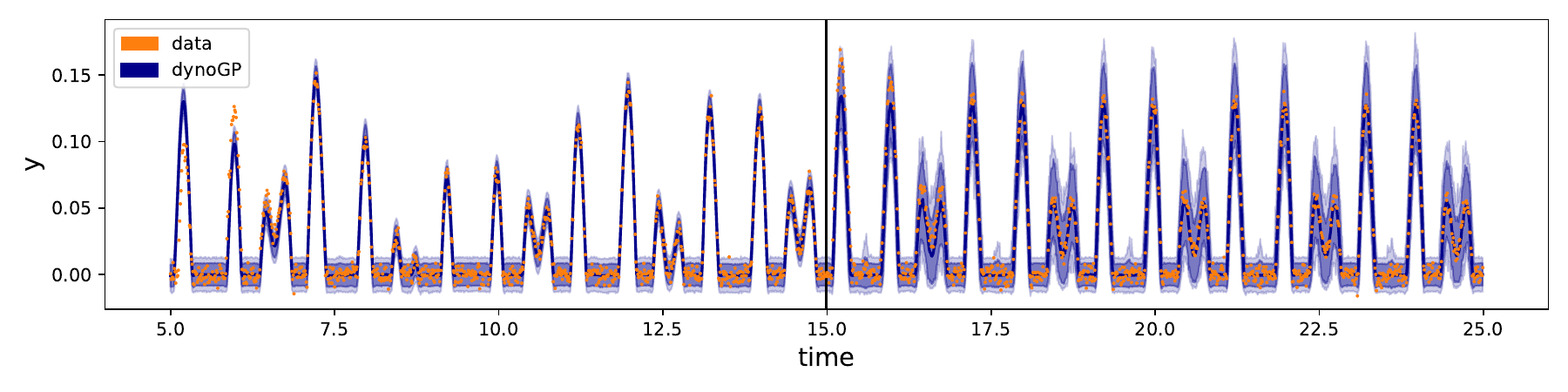}
\caption{Example: system identification with dynoGP. Top: training data (output signal). Center: dynoGP with only a dynamic layer; values on the left of the vertical bar represent the  last 1000 values of the training set, while those on the right correspond to the test set. Bottom: dynoGP with Wiener architecture. The last two plots show the posterior mean (blue line), 95\% credible interval (darker blue-region), 99.7\% credible interval (lighter blue-region).}
    \end{figure*}
\end{example}

\begin{remark}

DeepGPs often suffer from degenerate covariances \cite{damianou2013deep}, where each layer in the composition reduces the rank of the system. To address this issue, \cite{salimbeni2017doubly} proposed using an identity mean function in all intermediate GP layers. Our model does not appear to exhibit this issue. This is because the dynamic GP layer inherently has a linear (identity) mean, derived from the LTI system. For static GP, we use a constant mean function when the static layer is an intermediate/final layer and a linear mean function when the static layer is an input layer.
\end{remark}

The extension to other architectures such as Wiener-Hammerstein $g^{(2)}_t\big(f \big(g^{(1)}_t\big({\bf u}(t)\big)\big)\big)$ or Hammerstein-Wiener $f^{(2)} \big(g_t\big(f^{(1)}{\bf u}(t)\big)\big)\big)$ can be similarly defined.
So far, we have assumed a static or dynamic block for each layer, following a MISO  structure. However, this approach can be generalized to a MIMO (Multiple Input, Multiple Output) structure by incorporating multiple GPs per layer (of the same type, static or dynamic).  We will use this more general architectrures in the next section. Additionally, \textit{skip connections} can be easily introduced to further enhance the model's flexibility. 
We implemented \textit{dynoGP} in \textit{GPyTorch} \cite{gardner2018gpytorch}. Additional details for the selection of the inducing points,  the prior over the hyperparameters, and the link to the GitHub repository are provided in Appendix \ref{app:impl}.

\section{Numerical experiments} \label{Sec:examples}

In order to demonstrate the effectiveness of the proposed dynoGP architecture, we consider  the following case studies:
\begin{itemize}
    \item identification of a simulated Wiener system;
    \item identification of the Wiener-Hammerstein benchmark with process noise \cite{schoukens2017three};
    \item identification the coupled-electric drives benchmark \cite{wigren2017coupled};
    \item forecast of electricity demand \cite{godahewa2021australian}.
\end{itemize}
The performance of the estimated models is measured in terms of \emph{Root Mean Square Error} (RMSE), \emph{Mean Absolute Error} (MAE), and \emph{Continuous Ranked Probability Score} (CRPS), defined respectively by the following equations:
\begin{align}
    \text{RMSE} = & \sqrt{\frac{1}{N} \sum_{i=1}^N (y_i - \hat{y}_i)^2} \\
    \text{MAE} = & \frac{1}{N} \sum_{i=1}^N |y_i - \hat{y}_i| \\
    \text{CRPS} = & \frac{1}{N} \sum_{i=1}^N \int_{-\infty}^{\infty} (F_i(x) - H(x - y_i))^2 \, dx
\end{align}
where $y_i$ is the observed output at time stamp $i$, $\hat{y}_i$ is the predicted output, $N$ is the number of observations, $H$ is the Heaviside step function, and $F_i(x)$ is the cumulative distribution function associated with the prediction of the $i$-th output sample.
It is worth remarking that the RMSE and the MAE are computed using the mean of the estimated posterior, while the CRPS allows measuring the performance of the approach by evaluating the entire probabilistic output.  This metric is particularly useful for evaluating predictions that express uncertainty in their outputs, providing a  measure of forecast quality that accounts for both the entire probability distribution. In case of point estimate, the CRPS coincides with the MAE. In the following, we compute the CRPS by sampling from the posterior distribution. 

In the simulations, we compare the accuracy of dynoGP against dynoNet and GP-NARX. For dynoNet, we employ the same architecture and dimensions for the dynamic layer as dynoGP. For the static layer, we use a feedforward  network with one hidden layer
containing $32$ neurons. For dynoGP, we use a Matern 3/2 kernel in the static layer.
For the choice of architecture (interconnection of linear and static blocks), we consistently use a LIN-NONLIN-LIN structure, with variations only in the number of blocks. An exception is the first simulation, where we employed a Wiener architecture, as it was used to generate the data.

GP-NARX is a GP model that uses past inputs and measured outputs as covariates, with input and output lags of order $n_{lags}$. It is worth noting that, in GP-NARX, the testing phase involves iteratively constructing the covariate using past predicted outputs, as the measured outputs are unavailable during prediction.

\subsection{Simulated Wiener system with process noise}
In the first experiment, we simulated data from a Wiener system, which consists of an LTI system followed by the nonlinearity $(\cdot)^+$ (that is, positive part). We considered a setup similar to those in Examples \ref{ex:2} and \ref{ex:3}. Specifically, we used a two-dimensional input signal $\mathbf{u}(t) = [\sin(3\pi t), \sin(5\pi t)]$ and simulated an output signal $y(t)$ using an LTI system with a 5-dimensional state, followed by a quadratic nonlinearity.
The LTI system has a transition matrix defined as diagonal plus rank-one, $-\Lambda - {\bf v}{\bf v}^\top$, where the elements of $\Lambda$ and ${\bf v}$ were independently and uniformly generated in $[0,1]$. The remaining parameters, $C$, $B$, and $L$, were independently sampled from normal distributions with variances $1$, $1$, and $0.1$, respectively. The standard deviation $\varsigma$ of the measurement noise was set to 10\% of the standard deviation of the system's output $y(t)$. A sampling time of $\delta = 0.01$ was used, and data were generated over the interval $t \in [0, 25]$. For system identification, data from $t \leq 15$ were used. The remaining data for $t > 15$, generated without process noise, were used for testing the model. 
We performed 50 Monte Carlo simulations in total. To identify the system and predict the test data, we used a dynoGP with a Wiener architecture, as described in Section \ref{sec:dynoGP}. The dynamic GP layer was initialized with a system dimension of 10. For GP-NARX, we used input and output lags of order $n_{lags}=5$. 

\begin{figure}
\centering
\includegraphics[width=8cm]{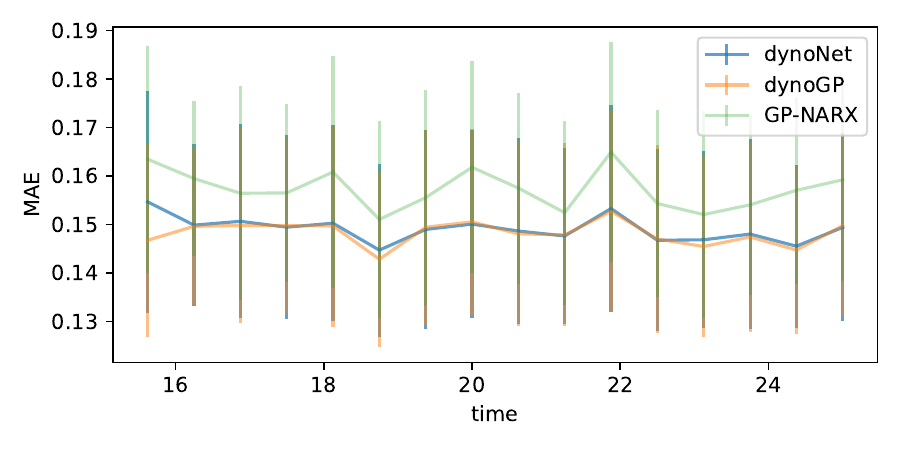}
\includegraphics[width=8cm]{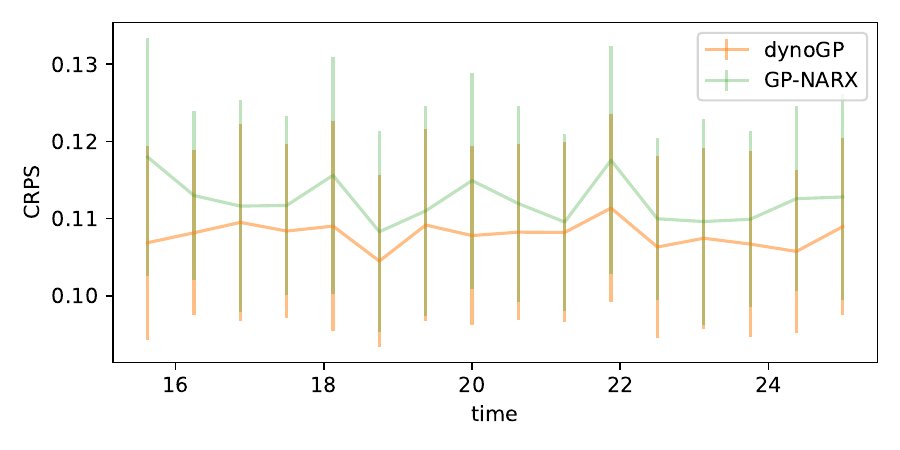}
\caption{Predictive MAE and CRPS for $15 \leq t \leq 25$ for dynoNet, dynoGP and GP-NARX for simulated data from a  Wiener system.}
\label{fig:maewiener}
\end{figure}

 Figure \ref{fig:maewiener}-top shows the MAE for the test data, showing the median and its standard deviation across the 50 Monte Carlo simulations as a function of time. The large error bars are caused by a high level of measurement noise. DynoGP and dynoNet provide a better performance than GP-NARX.  
 Note that, in the initial stages of the prediction task ($15<t<16.5$, corresponding to 150 values), the performance of dynoNet is  worse than dynoGP. This is theoretically expected, as dynoGP exhibits higher accuracy in short-term forecasting.  The reason is that dynoNet estimates only the deterministic part of the system, while the system's dynamics also depends on the process noise, which drives the output signal in the initial part of the testing task (fading memory effect). GP-NARX's higher MAE is due to usage of noisy covariates in the training data, which has an impact in the testing, where past simulated (instead of measured) outputs are used.    
 Figure \ref{fig:maewiener}-bottom shows the CRPS for the two probabilistic methods. We observe that dynoGP provides better probabilistic predictions than GP-NARX.

\subsection{Wiener-Hammerstein with process noise}
As a second case study, we consider the identification of the Wiener-Hammerstein benchmark system described  in \cite{schoukens2017three}. The system is characterised  by a static nonlinearity sandwiched between two Linear Time-Invariant (LTI) blocks. A process noise enters the system before the static nonlinearity, which makes the benchmark particularly challenging.

For identification via dynoGP, we consider the following Wiener-Hammerstein architecture: 

{  \centering
\includegraphics[width=9cm,trim={1cm 0 0 0},clip]{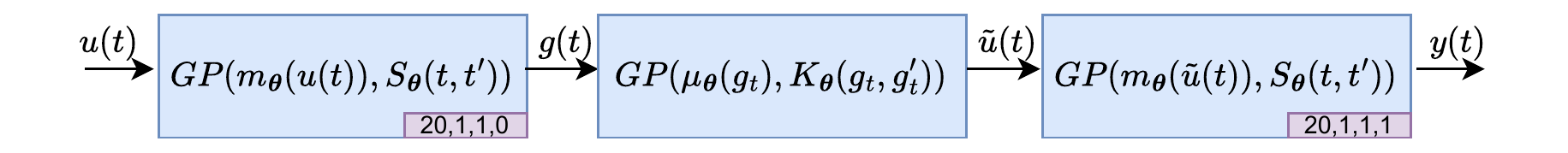}\vspace{-0.1cm} }

\begin{figure*}
\centering
\includegraphics[width=16cm]{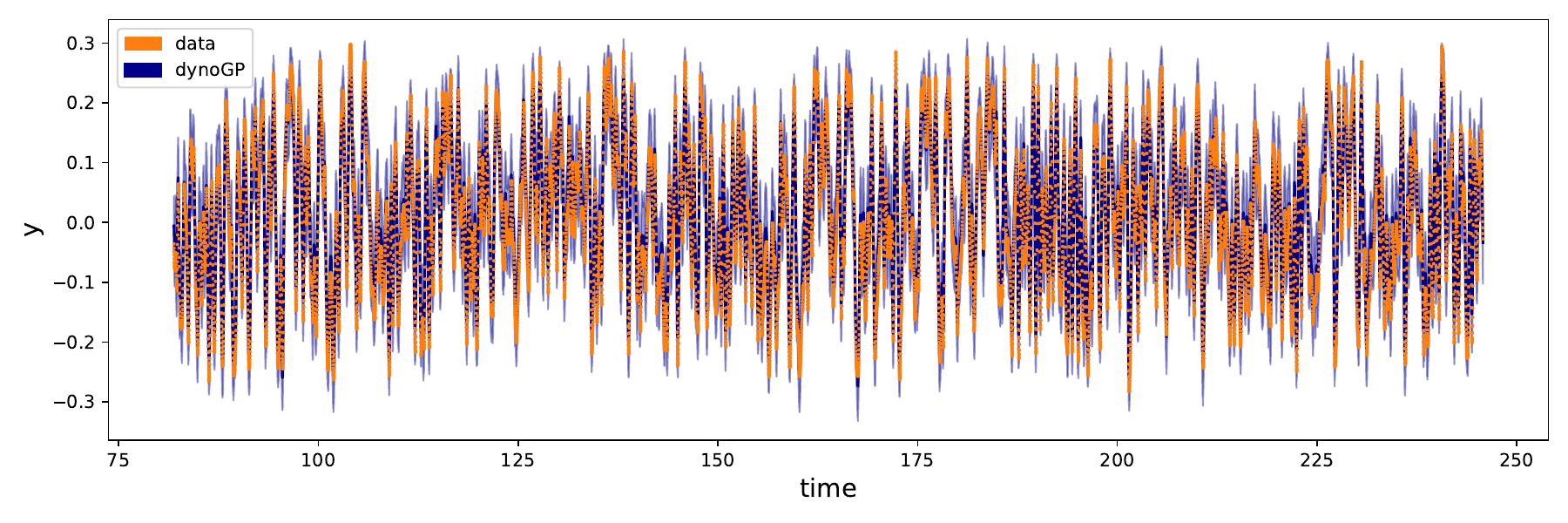}\vspace{-0.1cm}
\includegraphics[width=16cm]{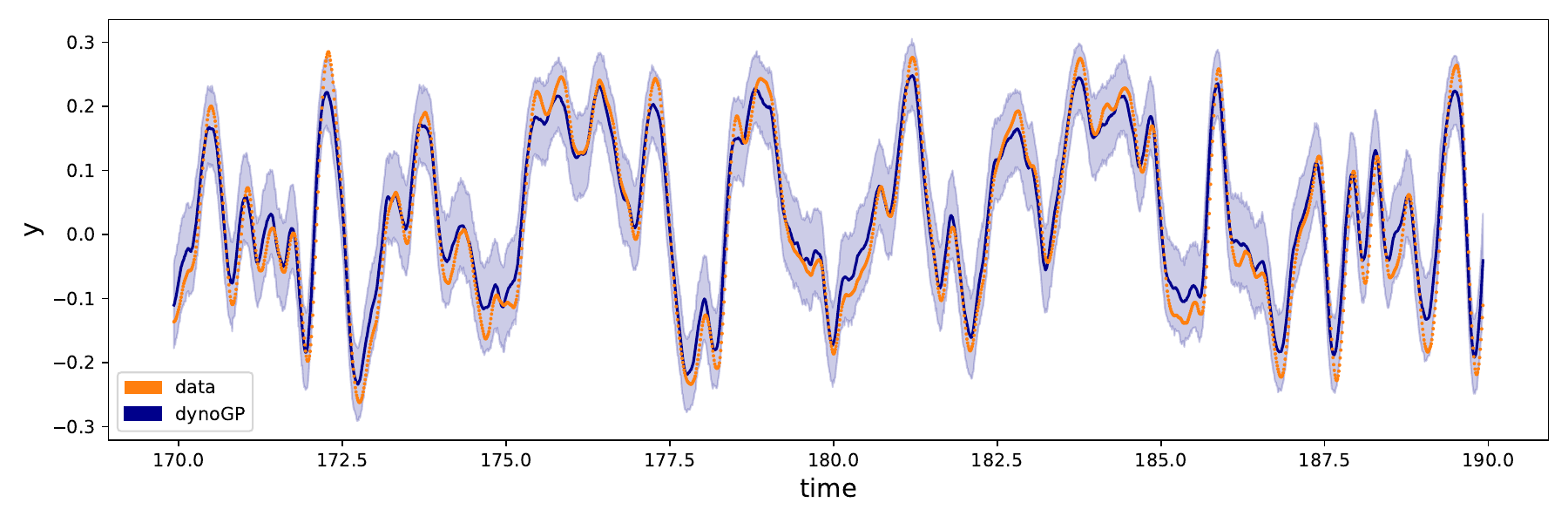}
\caption{Wiener-Hammerstein model with process noise. The output signal is in orange, the predictive posterior mean and 95\% credible interval for dynoGP are shown for all testing data (top) and for a subset of testing data (bottom).} 
\label{Fig_WH_processnoise}
\end{figure*}

Figure \ref{Fig_WH_processnoise}-top displays the output signal on the test dataset alongside the dynoGP posterior mean and the $95$\% credible interval obtained by dynoGP. For enhanced visualization, a subset of the test set is presented in Figure \ref{Fig_WH_processnoise}-bottom. A comparison with the dynoNet and GP-NARX architectures is detailed in Table \ref{tab:comparison}, focusing on the performance metrics RMSE, MAE, and CRPS. It is notable that the model's uncertainty is significantly lower compared to GP-NARX, as evidenced by the lower CRPS score. Additionally, both dynoGP and dynoNet architectures outperform GP-NARX in terms of RMSE and MAE, with dynoGP slightly outperforming dynoNet.

\subsection{Coupled Electric Drives}
As a third case study, we consider the identification of the coupled-electric drives benchmark \cite{wigren2017coupled}, which involves two electric motors driving a pulley via a flexible belt. The pulley is supported by a spring, introducing a lightly damped dynamic mode. The input is the sum of the  voltage applied to the motors, while the angular speed of the pulley is the output of interest. The dataset comprises 500 samples, with the first 400 used for training and the remaining 100  for testing.

We considered the following dynoGP architecture:

{ 
    \centering
\includegraphics[width=9cm,trim={1cm 0 0 0},clip]{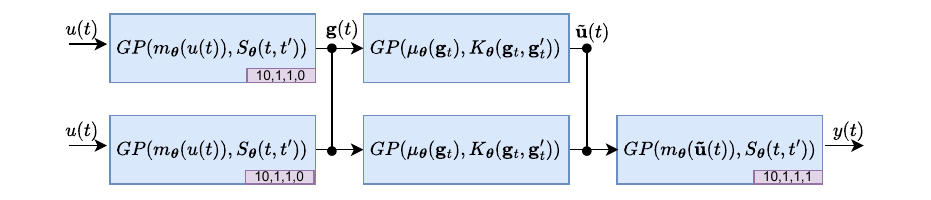}}

\begin{figure*}
    \centering
    \includegraphics[width=16cm]{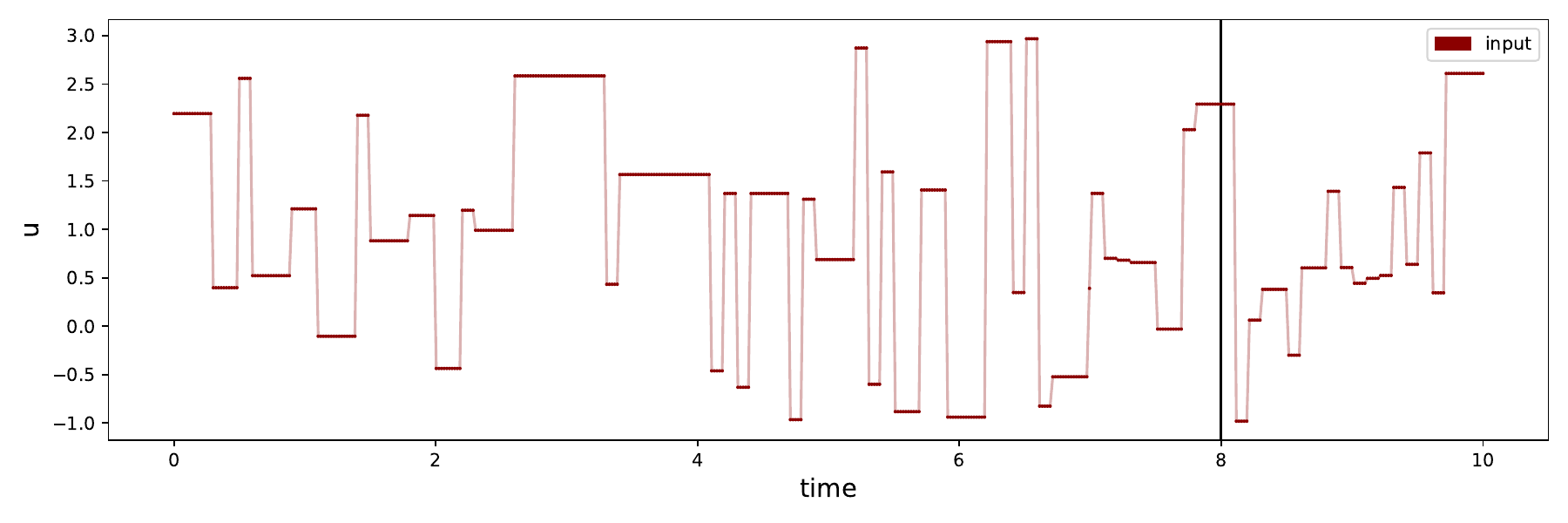}\vspace{-0.1cm}
    \includegraphics[width=16cm]{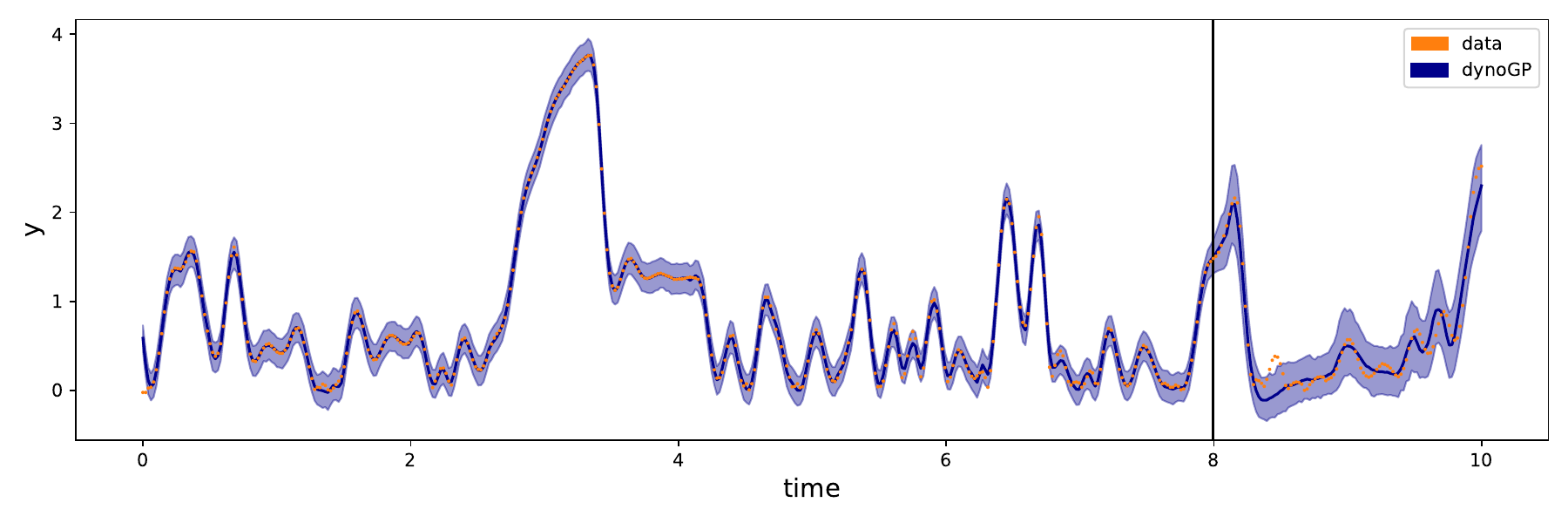}
\caption{CED:  The first plot shows the input signal. Values on the left of the vertical bar represent the  400 values of the training set, while those on the right correspond to the test set. The second plot shows the values of the output signal (in orange) and the predictive posterior mean and 95\% credible interval for dynoGP  in blue. } 
\label{fig:CEDresults}
\end{figure*}

Figure \ref{fig:CEDresults}-top illustrates the input signal, while Figure \ref{fig:CEDresults}-bottom presents the output signal (in orange) alongside the dynoGP posterior mean and the 95\% credible interval (in blue). The values on the left of the vertical bar correspond to the 400 data points in the training set, whereas those on the right represent the test set. It is evident that the model uncertainty is relatively high, which can be attributed to the small size of the training dataset (only 400 data points). Nevertheless, the model's uncertainty is notably lower compared to GP-NARX, as evidenced by the comparison of the CRPS of the two models in Table~\ref{tab:comparison}. 
The other metrics are also reported in Table \ref{tab:comparison}. Both dynoGP and dynoNet outperform the GP-NARX architecture, while  the performance of dynoNet is slightly better than dynoGP. However, we remind that  while dynoNet provides a point estimate, dynoGP provides additional information consisting of the probability distribution of the predicted output.

\subsection{Electricity demand forecast}
Finally, we considered the problem of forecasting electricity demand (in gigawatt) as a function of weather-temperature (in Celsius). The dataset comprises electricity demand in the state of Victoria, Australia, for the years 2014 and 2015. It includes two time series recorded at 30-minute intervals: electricity demand and maximum temperature.
Daily electricity demand is strongly influenced by temperature, with higher consumption observed on colder days due to heating and on hotter days due to air conditioning. This relationship exhibits nonlinearity due to threshold effects, emphasizing the necessity of employing both non-linear and dynamic models to capture the underlying patterns accurately. 
For this reason, we use an architecture which combines linear dynamical GPs and static GPs. In particular, we consider the following architecture:

{\centering
    \includegraphics[width=9cm]{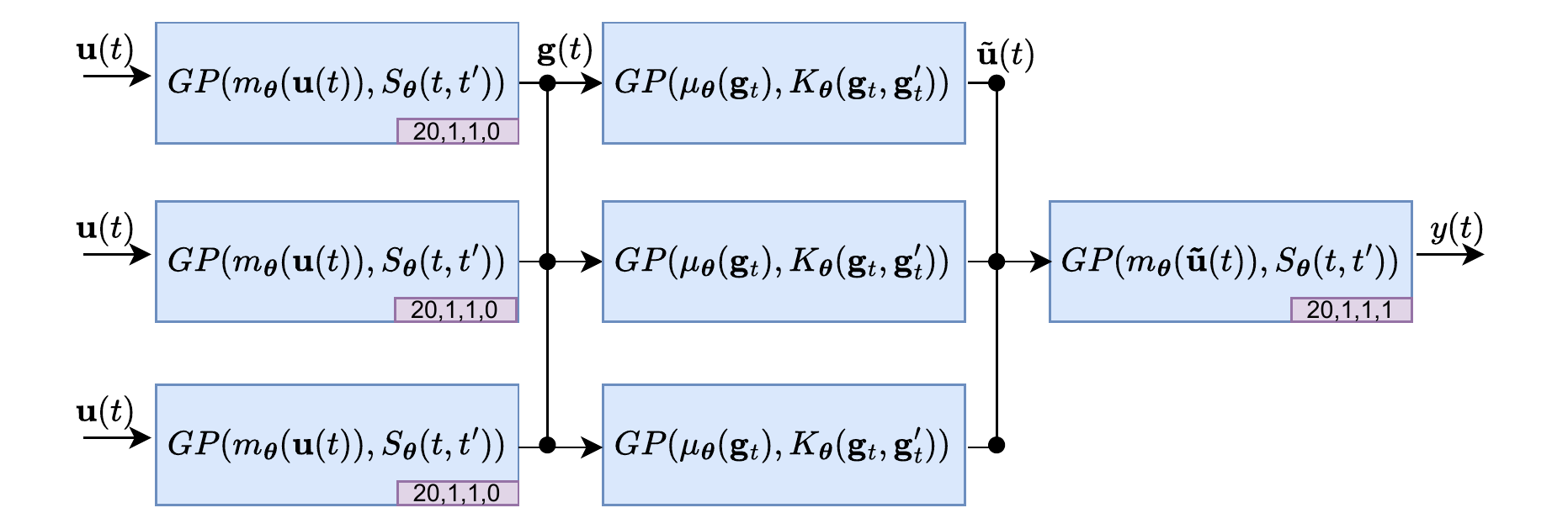}
}

For our analysis, we used 18 months of data for training (comprising 26,304 values) and 6 months for testing (8,736 values). The inputs to the model included maximum temperature, an indicator variable denoting whether a day was a public holiday, and Fourier harmonics:
$$
[\cos(2 \pi t m_i),\sin(2 \pi t m_i),\dots,\cos(2 \pi t k_i m_i),\sin(2 \pi t k_i m_i)]
$$
for $t=0,\delta, 2\delta,\dots$ and $\delta=1/(48\cdot 365)$ and three terms for the three periods: daily ($m_1=365$ , $k_1=4$), weekly ($m_2=52.14$ , $k_2=4$) and yearly ($m_2=1$ , $k_2=2$). In total, we included 22 input signals for forecasting electricity demand.
It is important to note that this setup is not entirely realistic, as one of the input, maximum temperature, is  unavailable for more than 15 days ahead, which is the typical weather forecasting horizon. Instead, our testing set includes predictions for up to 6 months ahead. In this section, our primary objective is to use this time series to learn the corresponding dynamical system and test dynoGP performance for long time horizon and in a scenario with many input signals.

\begin{figure*}
    \centering
            \includegraphics[width=16cm]{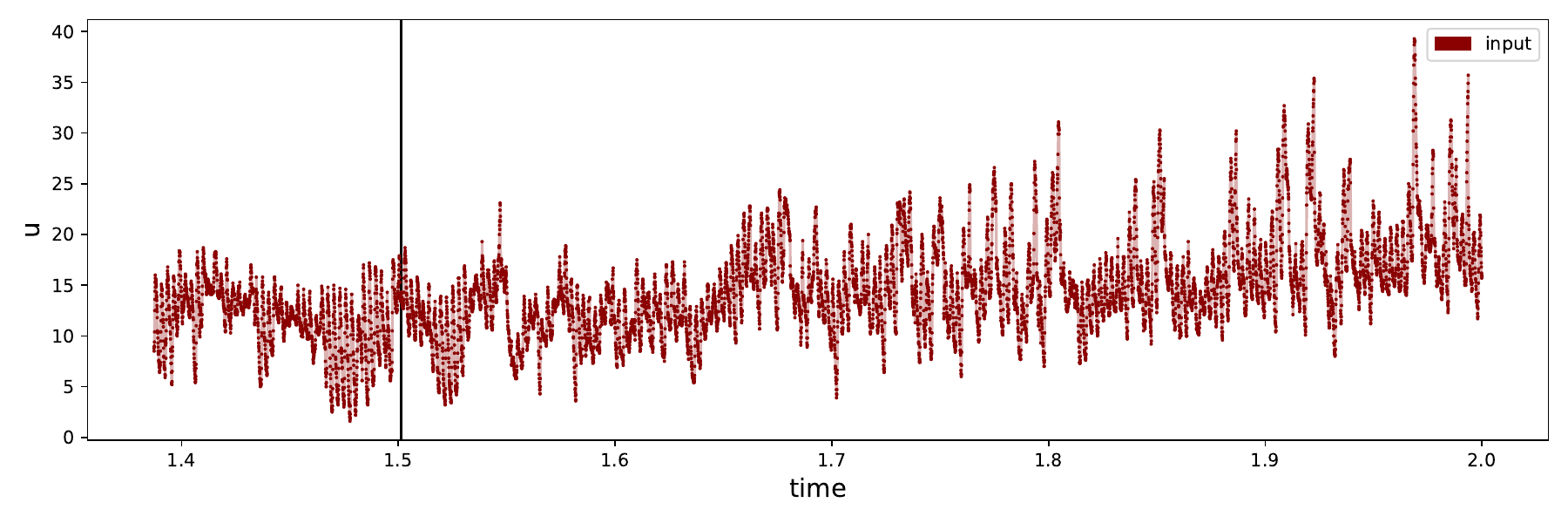}\vspace{-0.1cm}
        \includegraphics[width=16cm]{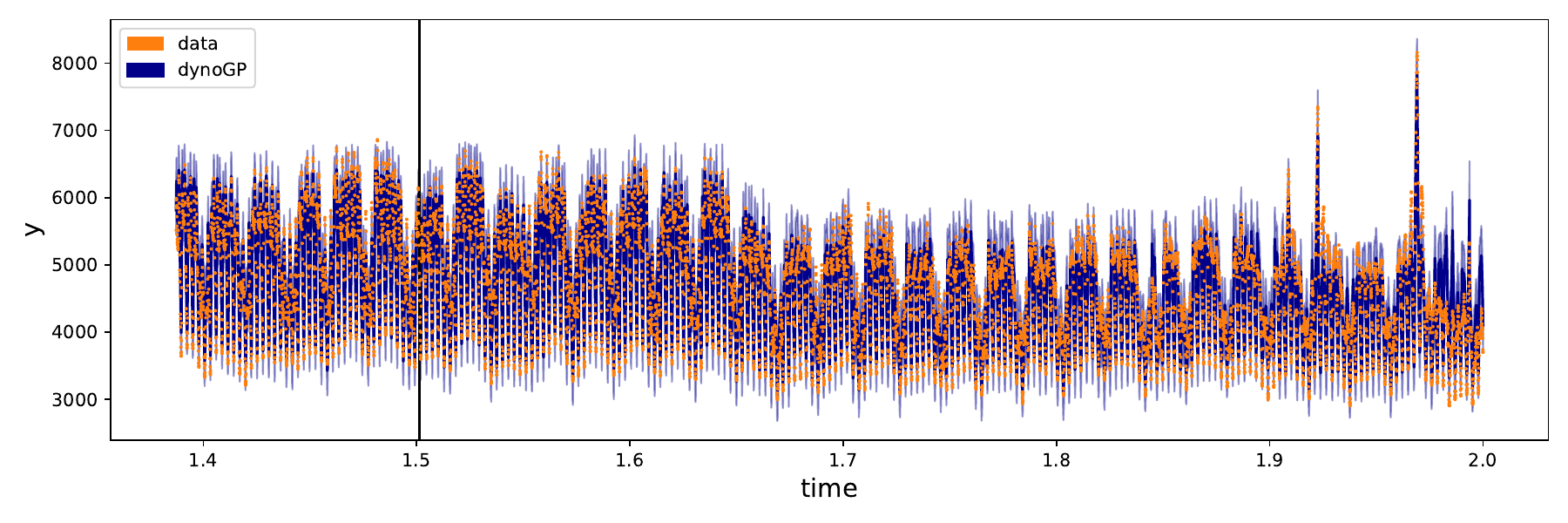}
                \includegraphics[width=16cm]{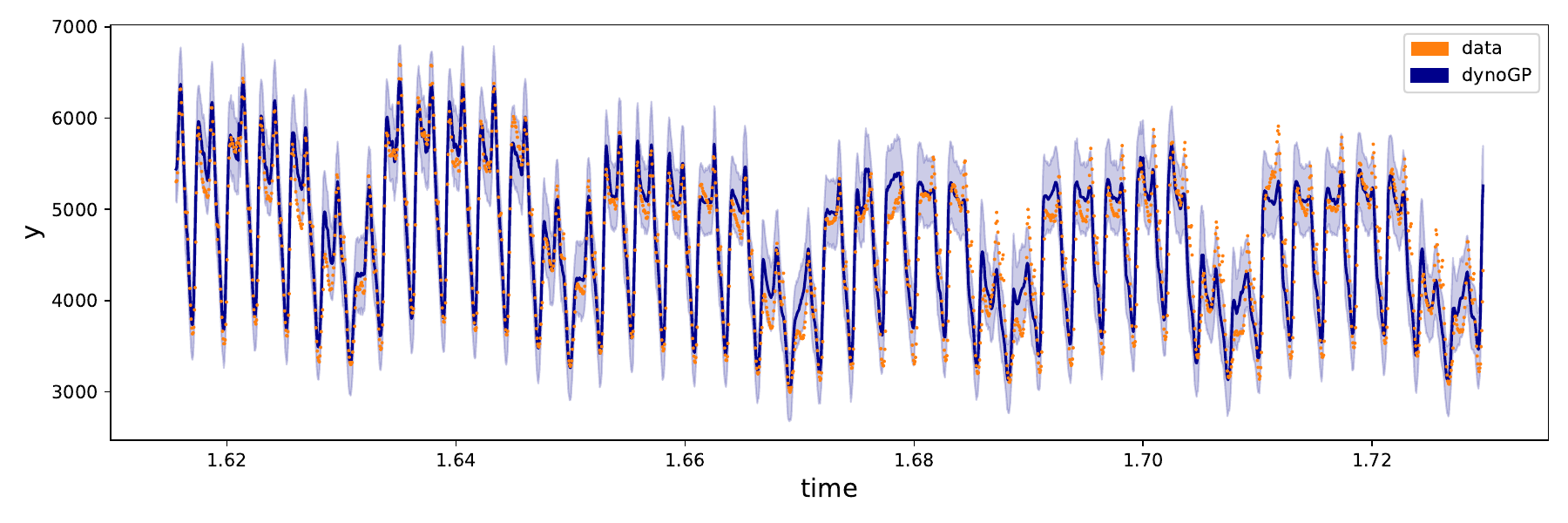}
                
\caption{Electricity demand dataset. The first plot shows the maximum temperature (input). Values on the left of the vertical bar represent the last 2,000 values of the training set, while those on the right correspond to the test set. The second plot includes  the corresponding values of the electricity demand (in orange) and shows the predictive posterior mean and 95\% credible interval for dynoGP. The last plot is a zoomed-in view for a subset of the prediction task. } \label{fig:ELEresults}
    \end{figure*}

Figure \ref{fig:ELEresults} (first and second plot) shows a subset of the input temperature and electricity demand time series. The values on the left of the vertical bar represent the last 2,000 observations of the training set, while those on the right correspond to the test set. The third  plot in Figure \ref{fig:ELEresults} displays the posterior mean and 95\% credible interval for  dynoGP (the third plot is a zoomed-in view). The results indicate that the model predicts electricity demand accurately even 6 months ahead. Notably, it performs exceptionally well in forecasting peak demands, which are crucial for ensuring the stability of the electricity grid. Note that, the model's performance diminishes in the last part of the testing set. This pattern is not evident from the temperature, suggesting the presence of other influencing inputs not captured by the model. 
As in the other benchmarks, a comparison with dynoNet and GP-NARX in terms of RMSE, MAE, and CRPS is also  provided in Table \ref{tab:comparison}. 

\begin{table*}[h!]
\centering
\begin{tabular}{@{}lccc|cc|ccc@{}}
\toprule  
\multirow{2}{*}{Datasets} & \multicolumn{3}{c|}{dynoGP}         & \multicolumn{2}{c|}{dynoNet}         & \multicolumn{3}{c}{GP-NARX}         \\ 
                          & RMSE          & MAE          & CRPS          & RMSE          & MAE                    & RMSE          & MAE          & CRPS          \\ \midrule
WH                 & 0.029         & 0.024        & 0.016        &     0.034     &   0.028           &     0.120     &   0.096   & 0.077        \\
CED                & 0.147         & 0.109        & 0.081        &     0.105     &   0.075            &      0.247    &   0.174   & 0.132          \\
ELE  & 291         & 215        & 197        &     297        & 235         &    376      & 273    & 258         \\

\bottomrule
\end{tabular}
\caption{Performance comparison of dynoGP, dynoNet, and GP-NARX in terms of RMSE, MAE, and CRPS metrics across the benchmark datasets: Wiener-Hammerstein (WH) with process noise;  Coupled Electric Drives (CED); electricity demand forecast (ELE).}
\label{tab:comparison}
\end{table*}

\subsection{Computational time}
Table \ref{tab:time} shows the per-batch computation time for dynoGP on an NVIDIA GeForce RTX 4060 GPU for the four problems discussed previously.  The differences in time are primarily due to the varying complexity of the architectures (number of linear and static blocks and number of inputs). The number of training iterations through the dataset is provided in Appendix \ref{app:impl}. The mini-batch implementation of dynoGP allows it to scale to thousands of observations.

\begin{table}[h!]
\centering
\begin{tabular}{l|c|c|}
\toprule  
\multirow{1}{*}{Datasets} & \multicolumn{1}{c|}{mini-batch size}         & \multicolumn{1}{c|}{time (sec)}  \\ \hline
Wiener & 375 & 0.7\\
WH   & 1024  &  3  \\
CED  & 400 &   1   \\
ELE  & 1053 &   6   \\
\bottomrule
\end{tabular}
\caption{Computational time per mini-batch.}
\label{tab:time}
\end{table}

\section{Conclusions}
The deep dynamic GP architecture (dynoGP) presented in this paper is based on a time-dependent GP process associated to stochastic linear time-invariant dynamical systems with a complex diagonal state-transition matrix. This setup facilitates an efficient and closed-form expression for the mean and covariance function of the dynamic GP. The linear time-invariant GP can be cascaded with other GP blocks, both static and dynamic, to form the dynoGP architecture, where the posterior distribution of the predicted output can be approximated through    stochastic variational inference.

As future work, we plan to explore three key aspects. First, we aim to investigate model selection for the dynoGP architecture by leveraging Pareto Smoothed Importance Sampling Leave-One-Out cross-validation (PSIS-LOO) \cite{vehtari2017practical}. Exact cross-validation requires re-fitting the model on different training sets, which can be computationally expensive. In contrast, PSIS-LOO can be efficiently computed using samples from the posterior distribution. Second, we intend to develop a principled approach for initialising the parameters of the dynamic and static GP layers. Our idea is to employ a hierarchical Bayesian framework, as discussed in \cite{corani2021time,Benavoli2021eb} for GP-based time-series forecasting. Additionally, we believe that this approach can be adapted to extend dynoGP for metalearning tasks. Metalearning involves training models to quickly adapt to new tasks using only a small amount of data, which aligns well with the flexibility and expressiveness of Gaussian processes. By leveraging the hierarchical Bayesian approach, we can encode prior knowledge about task distributions into the model, enabling it to generalise across tasks more effectively. This could make dynoGP particularly well-suited for applications where task-specific data is limited but shared structure across tasks can be exploited.
Third, we plan to apply dynoGP to various sequence modelling problems (e.g., images, audio, text, time-series), leveraging its similarity to structured state space sequence models  (like S4, Mamba, which have demonstrated strong performance on established benchmarks), while providing a full probabilistic output.

\bibliographystyle{plain}   
\bibliography{biblio}

\begin{thebibliography}{10}

\bibitem{andersson2019deep}
Carl Andersson, Ant{\^o}nio~H Ribeiro, Koen Tiels, Niklas Wahlstr{\"o}m, and
  Thomas~B Sch{\"o}n.
\newblock Deep convolutional networks in system identification.
\newblock In {\em 2019 IEEE 58th Conference on Decision and Control (CDC)},
  pages 3670--3676. IEEE, 2019.

\bibitem{bai2018empirical}
Shaojie Bai, J~Zico Kolter, and Vladlen Koltun.
\newblock An empirical evaluation of generic convolutional and recurrent
  networks for sequence modeling.
\newblock {\em arXiv preprint arXiv:1803.01271}, 2018.

\bibitem{bauer2016understanding}
Matthias Bauer, Mark van~der Wilk, and Carl~Edward Rasmussen.
\newblock Understanding probabilistic sparse {G}aussian process approximations.
\newblock In {\em Advances in neural information processing systems}, pages
  1533--1541, 2016.

\bibitem{beintema2021nonlinear}
Gerben~I Beintema, Roland T{\'o}th, and Maarten Schoukens.
\newblock Nonlinear state-space identification using deep encoder networks.
\newblock In {\em Learning for Dynamics and Control}, pages 241--250. PMLR,
  2021.

\bibitem{benavoli2020skew}
Alessio Benavoli, Dario Azzimonti, and Dario Piga.
\newblock Skew gaussian processes for classification.
\newblock {\em Machine Learning}, 109(9):1877--1902, 2020.

\bibitem{benavoli2021}
Alessio Benavoli, Dario Azzimonti, and Dario Piga.
\newblock {A unified framework for closed-form nonparametric regression,
  classification, preference and mixed problems with Skew Gaussian Processes}.
\newblock {\em Machine Learning}, pages 1--39, 2021.

\bibitem{benavoli2021preferential}
Alessio Benavoli, Dario Azzimonti, and Dario Piga.
\newblock Preferential bayesian optimisation with skew gaussian processes.
\newblock In {\em Proceedings of the Genetic and Evolutionary Computation
  Conference Companion}, pages 1842--1850, 2021.

\bibitem{Benavoli2021eb}
Alessio Benavoli and Giorgio Corani.
\newblock State space approximation of gaussian processes for time series
  forecasting.
\newblock pages 21--35, Cham, 2021. Springer International Publishing.

\bibitem{benavoli2016state}
Alessio Benavoli and Marco Zaffalon.
\newblock State space representation of non-stationary gaussian processes.
\newblock {\em arXiv preprint arXiv:1601.01544}, 2016.

\bibitem{boyd1985fading}
Stephen Boyd and Leon Chua.
\newblock Fading memory and the problem of approximating nonlinear operators
  with volterra series.
\newblock {\em IEEE Transactions on circuits and systems}, 32(11):1150--1161,
  1985.

\bibitem{bui2016deep}
Thang Bui, Daniel Hern{\'a}ndez-Lobato, Jose Hernandez-Lobato, Yingzhen Li, and
  Richard Turner.
\newblock Deep gaussian processes for regression using approximate expectation
  propagation.
\newblock In {\em International conference on machine learning}, pages
  1472--1481. PMLR, 2016.

\bibitem{carron2016machine}
Andrea Carron, Marco Todescato, Ruggero Carli, Luca Schenato, and Gianluigi
  Pillonetto.
\newblock Machine learning meets kalman filtering.
\newblock In {\em 2016 IEEE 55th conference on decision and control (CDC)},
  pages 4594--4599. IEEE, 2016.

\bibitem{carvalho2009handling}
Carlos~M Carvalho, Nicholas~G Polson, and James~G Scott.
\newblock Handling sparsity via the horseshoe.
\newblock In {\em Artificial intelligence and statistics}, pages 73--80. PMLR,
  2009.

\bibitem{chen1990non}
S.~Chen, S.~A. Billings, and P.~M. Grant.
\newblock Non-linear system identification using neural networks.
\newblock {\em International journal of control}, 51(6):1191--1214, 1990.

\bibitem{NIPS2009_5751ec3e}
Youngmin Cho and Lawrence Saul.
\newblock Kernel methods for deep learning.
\newblock In Y.~Bengio, D.~Schuurmans, J.~Lafferty, C.~Williams, and
  A.~Culotta, editors, {\em Advances in Neural Information Processing Systems},
  volume~22. Curran Associates, Inc., 2009.

\bibitem{corani2021time}
Giorgio Corani, Alessio Benavoli, and Marco Zaffalon.
\newblock Time series forecasting with gaussian processes needs priors.
\newblock In {\em Joint European Conference on Machine Learning and Knowledge
  Discovery in Databases}, pages 103--117. Springer, 2021.

\bibitem{damianou2013deep}
Andreas Damianou and Neil~D Lawrence.
\newblock Deep gaussian processes.
\newblock In {\em Artificial intelligence and statistics}, pages 207--215.
  PMLR, 2013.

\bibitem{forgione2021dynonet}
M.~Forgione and D.~Piga.
\newblock \emph{dyno{Net}}: A neural network architecture for learning
  dynamical systems.
\newblock {\em International Journal of Adaptive Control and Signal
  Processing}, 35(4):612--626, 2021.

\bibitem{forgione2020model}
Marco Forgione and Dario Piga.
\newblock Model structures and fitting criteria for system identification with
  neural networks.
\newblock In {\em 2020 IEEE 14th International Conference on Application of
  Information and Communication Technologies (AICT)}, pages 1--6. IEEE, 2020.

\bibitem{gardner2018gpytorch}
Jacob Gardner, Geoff Pleiss, Kilian~Q Weinberger, David Bindel, and Andrew~G
  Wilson.
\newblock Gpytorch: Blackbox matrix-matrix gaussian process inference with gpu
  acceleration.
\newblock {\em Advances in neural information processing systems}, 31, 2018.

\bibitem{gedon2021deepssm}
Daniel Gedon, Niklas Wahlstr{\"o}m, Thomas~B. Sch{\"o}n, and Lennart Ljung.
\newblock Deep state space models for nonlinear system identification.
\newblock In {\em Proceedings of the 19th IFAC Symposium on System
  Identification (SYSID)}, July 2021.
\newblock online.

\bibitem{gibbs2000variational}
Mark~N Gibbs and David~JC MacKay.
\newblock {Variational Gaussian process classifiers}.
\newblock {\em IEEE Transactions on Neural Networks}, 11(6):1458--1464, 2000.

\bibitem{godahewa2021australian}
R.~Godahewa, C.~Bergmeir, G.~Webb, R.~Hyndman, and P.~Montero-Manso.
\newblock Australian electricity demand dataset, 2021.

\bibitem{gu2020hippo}
Albert Gu, Tri Dao, Stefano Ermon, Atri Rudra, and Christopher R{\'e}.
\newblock Hippo: Recurrent memory with optimal polynomial projections.
\newblock {\em Advances in neural information processing systems},
  33:1474--1487, 2020.

\bibitem{gu2022parameterization}
Albert Gu, Karan Goel, Ankit Gupta, and Christopher R{\'e}.
\newblock On the parameterization and initialization of diagonal state space
  models.
\newblock {\em Advances in Neural Information Processing Systems},
  35:35971--35983, 2022.

\bibitem{gupta2022diagonal}
Ankit Gupta, Albert Gu, and Jonathan Berant.
\newblock Diagonal state spaces are as effective as structured state spaces.
\newblock {\em Advances in Neural Information Processing Systems},
  35:22982--22994, 2022.

\bibitem{hartikainen2010kalman}
Jouni Hartikainen and Simo S{\"a}rkk{\"a}.
\newblock Kalman filtering and smoothing solutions to temporal gaussian process
  regression models.
\newblock In {\em 2010 IEEE international workshop on machine learning for
  signal processing}, pages 379--384. IEEE, 2010.

\bibitem{Hensman2013}
James Hensman, Nicol\`{o} Fusi, and Neil~D. Lawrence.
\newblock Gaussian processes for big data.
\newblock In {\em Proceedings of the Twenty-Ninth Conference on Uncertainty in
  Artificial Intelligence}, UAI'13, pages 282--290, Arlington, Virginia, USA,
  2013. AUAI Press.

\bibitem{hernandez2016scalable}
Daniel Hernandez-Lobato and Jose~Miguel Hernandez-Lobato.
\newblock Scalable gaussian process classification via expectation propagation.
\newblock In {\em Proceedings of the 19th International Conference on
  Artificial Intelligence and Statistics}, page 168–176. PMLR, 2016.

\bibitem{Lee2018}
Jaehoon Lee, Yasaman Bahri, Roman Novak, Samuel~S. Schoenholz, Jeffrey
  Pennington, and Jascha~Narain Sohl-Dickstein.
\newblock Deep neural networks as gaussian processes.
\newblock In {\em International Conference on Learning Representations}. Curran
  Associates, Inc., 2018.

\bibitem{loper2021general}
Jackson Loper, David Blei, John~P Cunningham, and Liam Paninski.
\newblock A general linear-time inference method for gaussian processes on one
  dimension.
\newblock {\em Journal of Machine Learning Research}, 22(234):1--36, 2021.

\bibitem{mackay1996bayesian}
David~JC MacKay.
\newblock Bayesian methods for backpropagation networks.
\newblock In {\em Models of neural networks III}, pages 211--254. Springer,
  1996.

\bibitem{masti2021learning}
Daniele Masti and Alberto Bemporad.
\newblock Learning nonlinear state--space models using autoencoders.
\newblock {\em Automatica}, 129:109666, 2021.

\bibitem{minka2001family}
Thomas~Peter Minka.
\newblock {\em A family of algorithms for approximate Bayesian inference}.
\newblock PhD thesis, Massachusetts Institute of Technology, 2001.

\bibitem{narendra1990identification}
Kumpati~S Narendra and Kannan Parthasarathy.
\newblock Identification and control of dynamical systems using neural
  networks.
\newblock {\em IEEE Transactions on neural networks}, 1(1):4--27, 1990.

\bibitem{o1978curve}
Anthony O'Hagan.
\newblock Curve fitting and optimal design for prediction.
\newblock {\em Journal of the Royal Statistical Society: Series B
  (Methodological)}, 40(1):1--24, 1978.

\bibitem{opper2009variational}
Manfred Opper and C{\'e}dric Archambeau.
\newblock The variational gaussian approximation revisited.
\newblock {\em Neural computation}, 21(3):786--792, 2009.

\bibitem{orvieto2023resurrecting}
Antonio Orvieto, Samuel~L Smith, Albert Gu, Anushan Fernando, Caglar Gulcehre,
  Razvan Pascanu, and Soham De.
\newblock Resurrecting recurrent neural networks for long sequences.
\newblock In {\em International Conference on Machine Learning}, pages
  26670--26698. PMLR, 2023.

\bibitem{pillonetto2025deep}
Gianluigi Pillonetto, Aleksandr Aravkin, Daniel Gedon, Lennart Ljung,
  Ant{\^o}nio~H Ribeiro, and Thomas~B Sch{\"o}n.
\newblock Deep networks for system identification: a survey.
\newblock {\em Automatica}, 171:111907, 2025.

\bibitem{quinonero2005unifying}
Joaquin Qui{\~n}onero-Candela and Carl~Edward Rasmussen.
\newblock A unifying view of sparse approximate {G}aussian process regression.
\newblock {\em Journal of Machine Learning Research}, 6(Dec):1939--1959, 2005.

\bibitem{rasmussen2006gaussian}
Carl~Edward Rasmussen and Christopher~KI Williams.
\newblock {\em {Gaussian processes for machine learning}}.
\newblock MIT press Cambridge, MA, 2006.

\bibitem{salimbeni2017doubly}
Hugh Salimbeni and Marc Deisenroth.
\newblock Doubly stochastic variational inference for deep gaussian processes.
\newblock {\em Advances in neural information processing systems}, 30, 2017.

\bibitem{sarkka2019applied}
Simo S{\"a}rkk{\"a} and Arno Solin.
\newblock {\em Applied stochastic differential equations}, volume~10.
\newblock Cambridge University Press, 2019.

\bibitem{schetzen2006volterra}
Martin Schetzen.
\newblock {\em The Volterra and Wiener theories of nonlinear systems}.
\newblock Krieger Publishing Co., Inc., 2006.

\bibitem{schoukens2017three}
M.~Schoukens and J.P. No{\"e}l.
\newblock Three benchmarks addressing open challenges in nonlinear system
  identification.
\newblock In {\em Proceedings of the 20th World Congress of the International
  Federation of Automatic Control}, pages 448--453, Toulouse, France, 2017.

\bibitem{SCHURCH2020}
Manuel Schuerch, Dario Azzimonti, Alessio Benavoli, and Marco Zaffalon.
\newblock {Recursive estimation for sparse Gaussian process regression}.
\newblock {\em Automatica}, 120:109--127, 2020.

\bibitem{schuch2023correlated}
Manuel Schuerch, Dario Azzimonti, Alessio Benavoli, and Marco Zaffalon.
\newblock Correlated product of experts for sparse gaussian process regression.
\newblock {\em Machine Learning}, 2023.

\bibitem{snelson2006sparse}
Edward Snelson and Zoubin Ghahramani.
\newblock {Sparse Gaussian processes using pseudo-inputs}.
\newblock In {\em Advances in neural information processing systems}, pages
  1257--1264, 2006.

\bibitem{solin2014explicit}
Arno Solin and Simo S{\"a}rkk{\"a}.
\newblock Explicit link between periodic covariance functions and state space
  models.
\newblock In {\em Artificial Intelligence and Statistics}, pages 904--912.
  PMLR, 2014.

\bibitem{pmlrv5titsias09a}
Michalis Titsias.
\newblock Variational learning of inducing variables in sparse {G}aussian
  processes.
\newblock In David van Dyk and Max Welling, editors, {\em Proceedings of the
  Twelth International Conference on Artificial Intelligence and Statistics},
  volume~5 of {\em Proceedings of Machine Learning Research}, pages 567--574,
  Hilton Clearwater Beach Resort, Clearwater Beach, Florida USA, 5 2009. PMLR.

\bibitem{vehtari2017practical}
Aki Vehtari, Andrew Gelman, and Jonah Gabry.
\newblock Practical bayesian model evaluation using leave-one-out
  cross-validation and waic.
\newblock {\em Statistics and computing}, 27:1413--1432, 2017.

\bibitem{wang2017system}
Yizhou Wang and Alejandro Gonzalez.
\newblock System identification using deep learning: State-of-the-art and
  future research directions.
\newblock {\em Annual Reviews in Control}, 43:201--213, 2017.

\bibitem{wigren2017coupled}
Torbj{\"o}rn Wigren and Maarten Schoukens.
\newblock Coupled electric drives data set and reference models.
\newblock Technical Reports 2017-024, Department of Information Technology,
  Uppsala University, Uppsala, Sweden, December 2017.

\bibitem{williams1996computing}
Christopher Williams.
\newblock Computing with infinite networks.
\newblock {\em Advances in neural information processing systems}, 9, 1996.

\bibitem{williams1998bayesian}
Christopher~KI Williams and David Barber.
\newblock Bayesian classification with gaussian processes.
\newblock {\em IEEE Transactions on Pattern Analysis and Machine Intelligence},
  20(12):1342--1351, 1998.

\bibitem{wray1994neural}
J~Wray and GGR Green.
\newblock Neural networks for nonlinear model structures.
\newblock {\em Control Systems Technology, IEEE Transactions on}, 2(1):31--41,
  1994.

\end{thebibliography}

\appendix 
\section*{Appendix}
\section{Proofs}
\label{app:proofs}
\textbf{Proof of Lemma \ref{lem:1}:}
The Lyapunov equation is 
\begin{equation}
\begin{aligned}
&A\Sigma_{\infty}+\Sigma_{\infty} A^H+L L^H\\
=& -\begin{bmatrix}
\lambda   & 0 \\
0  &  \lambda^\dagger 
\end{bmatrix}\begin{bmatrix}
\frac{LL^H}{\lambda+\lambda^\dagger}  & \frac{LL^\top}{2\lambda} \\
\frac{L^\dagger L^H}{2\lambda^\dagger}  & \frac{L^\dagger L^\top}{\lambda+\lambda^\dagger} \\
\end{bmatrix}-\begin{bmatrix}
\frac{LL^H}{\lambda+\lambda^\dagger}  & \frac{LL^\top}{2\lambda} \\
\frac{L^\dagger L^H}{2\lambda^\dagger}  & \frac{L^\dagger L^\top}{\lambda+\lambda^\dagger} \\
\end{bmatrix}\begin{bmatrix}
\lambda^\dagger   & 0 \\
0  &  \lambda 
\end{bmatrix}\\
&+\begin{bmatrix}
LL^H & LL^\top\\
L^\dagger L^H & L^\dagger L^\top 
\end{bmatrix}=0.
\end{aligned}
\end{equation}
~\\

\textbf{Proof of Proposition \ref{prop:1}:} For $t\leq t'$, the kernel function defined in \eqref{eq:covcompelx} is equal to:
\begin{equation}
\begin{aligned}
&S_{\boldsymbol{\theta}}(t,t')\\
=&-\begin{bmatrix}
c  & c^\dagger\\
\end{bmatrix}\begin{bmatrix}
\frac{LL^H}{\lambda+\lambda^\dagger}  & \frac{LL^\top}{2\lambda} \\
\frac{L^\dagger L^H}{2\lambda^\dagger}  & \frac{L^\dagger L^\top}{\lambda+\lambda^\dagger} \\
\end{bmatrix}\begin{bmatrix}
e^{\lambda^\dagger (t'-t)} & 0\\
0 & e^{\lambda (t'-t)}\end{bmatrix} \begin{bmatrix}
c^\dagger \\
c
\end{bmatrix}\\
=&-e^{\lambda^\dagger (t'-t)}\left(\tfrac{cLL^Hc^\dagger}{\lambda+\lambda^\dagger}  +\tfrac{c^\dagger L^\dagger L^Hc^\dagger }{2\lambda^\dagger} \right)\\
&~-e^{\lambda (t'-t)}\left(\tfrac{cLL^\top c}{2\lambda}  +\tfrac{c^\dagger L^\dagger L^\top c }{\lambda+\lambda^\dagger} \right).
\end{aligned}
\end{equation}
 For $t'< t$, the kernel function  is equal to:
\begin{equation}
\begin{aligned}
&S_{\boldsymbol{\theta}}(t,t')\\
=&-\begin{bmatrix}
c  & c^\dagger\\
\end{bmatrix}\begin{bmatrix}
e^{\lambda^\dagger (t-t')} & 0\\
0 & e^{\lambda (t-t')}\end{bmatrix} \begin{bmatrix}
\frac{LL^H}{\lambda+\lambda^\dagger}  & \frac{LL^\top}{2\lambda} \\
\frac{L^\dagger L^H}{2\lambda^\dagger}  & \frac{L^\dagger L^\top}{\lambda+\lambda^\dagger} \\
\end{bmatrix}\begin{bmatrix}
c^\dagger \\
c
\end{bmatrix}\\
=&-e^{\lambda^\dagger  (t-t')}\tfrac{cLL^Hc^\dagger}{\lambda+\lambda^\dagger}  -e^{\lambda  (t-t')}\tfrac{c^\dagger L^\dagger L^Hc^\dagger }{2\lambda^\dagger} \\
&-e^{\lambda^\dagger   (t-t')}\tfrac{cLL^\top c}{2\lambda}  -e^{\lambda  (t-t')}\tfrac{c^\dagger L^\dagger L^\top c }{\lambda+\lambda^\dagger}.
\end{aligned}
\end{equation}

\textbf{Proof of Proposition \ref{prop:2}:} The mean function of the system

\begin{equation}
\label{eq:complexsys1proof}
\begin{aligned}
d x_{c}(t) &= 
\overbrace{
 \begin{bmatrix}
\lambda   \\ 
\end{bmatrix}
}^{A_{c}}
x_{c}(t)dt + 
\overbrace{
\begin{bmatrix}
B
\end{bmatrix}
}^{B_{c}}
{\bf u}(t)dt \\
y(t) & = 2\Re\left(
\overbrace{\begin{bmatrix}c  \end{bmatrix}}^{C_{c}} x_{c}(t)\right)+D{\bf u}(t).
\end{aligned}
\end{equation}
 is:
\begin{equation}
\begin{aligned}
&{m}_{\boldsymbol{\theta}}(k\delta)\\
&=2\Re\left(ce^{\lambda k\delta}{\tilde m}(0)+ \sum_{i=1}^{k} \tfrac{c}{\lambda}e^{\lambda (i-1)\delta} (e^{\lambda \delta}-1)B\uu(i\delta)\right)\\
&+ D \uu(k\delta).
\end{aligned}
\end{equation}
Now consider the system in \eqref{eq:complexsys}, its mean function is:
\begin{equation}
\begin{aligned}
&m(k\delta)- D \uu(k\delta)\\&=C\left(\bar{A}^{k}{\bf m}(0)+ \sum_{i=1}^{k} \bar{A}^{i-1} \bar{B}\uu(i\delta)\right), \\
&=\begin{bmatrix}
c & c^\dagger
\end{bmatrix}\Bigg(\begin{bmatrix}
e^{\lambda \delta} & 0\\
0 & e^{\lambda^\dagger \delta} \\
\end{bmatrix}{\bf m}(0)+\sum_{i=1}^{k} \begin{bmatrix}
e^{\lambda \delta (i-1)} & 0\\
0 & e^{\lambda^\dagger \delta(i-1)} \\
\end{bmatrix}\\
&~~~~~~~~~\cdot\begin{bmatrix}
\frac{1}{\lambda} & 0\\
0 & \frac{1}{\lambda^\dagger} \\
\end{bmatrix}\begin{bmatrix}
e^{\lambda \delta }-1 & 0\\
0 & e^{\lambda^\dagger \delta}-1 \\
\end{bmatrix}\begin{bmatrix}
 B\\
 B^\dagger \\
\end{bmatrix}\uu(i\delta)\Bigg)\\
&=ce^{\lambda \delta}\tilde{m}(0)+c^\dagger e^{\lambda^\dagger \delta}\tilde{m}^\dagger(0)+\sum_{i=1}^{k}\Big(
\frac{c}{\lambda} e^{\lambda \delta (i-1)}(e^{\lambda \delta }-1) B\\
&+ \frac{c^\dagger}{\lambda^\dagger}e^{\lambda^\dagger \delta(i-1)}(e^{\lambda^\dagger \delta }-1) B^\dagger \Big)\uu(i\delta)\\
&=2\Re\Bigg(ce^{\lambda \delta}\tilde{m}(0)+\sum_{i=1}^{k}
\frac{c}{\lambda} e^{\lambda \delta (i-1)}(e^{\lambda \delta }-1) B \uu(i\delta)\Bigg)
\end{aligned}
\end{equation}
where ${\tilde m}(0)$ denotes the first component of ${\bf m}(0)$.\\

\textbf{Proof of Proposition \ref{prop:3}:} The result follows from Proposition \ref{prop:2} via block-independent stacking of the dynamical system and additive composition of the measurement equation.

\section{Details of the stochastic variational inference}
\label{app:svi}
We apply stochastic variational
inference (SVI) \cite{pmlrv5titsias09a,Hensman2013,salimbeni2017doubly}. In particular, we will consider a Wiener architecture to illustrate the SVI derivation. We start with the joint probability density augmented with inducing points for each GP block:
\begin{equation}
\scalebox{0.95}{$
\begin{aligned}
&p({\bf y},{\bf f},\Upsilon_f,{\bf g},\Upsilon_g|Z_f,Z_g,{\bf u})\\
&=\underbrace{p({\bf y}|{\bf f})}_{\text{likelihood}}\underbrace{p({\bf f},\Upsilon_f|Z_f,{\bf g})p(\Upsilon_f|Z_f)}_{\text{GP prior on $f$}}\underbrace{p({\bf g},\Upsilon_g|Z_g,{\bf u})p(\Upsilon_g|Z_g)}_{\text{GP prior on $g$}}
\end{aligned}$}
\end{equation}
where ${\bf y}=[y(\delta),\dots,y(n\delta)]^\top$, $Z_f,\Upsilon_f$ are the vector of inducing points for the static GP on the  function $f$,\\
${\bf f}=[f \big(g_{\delta}\big({\bf u}(\delta)\big)\big),\dots,f \big(g_{n\delta}\big({\bf u}(n\delta)\big)]^\top$,  $Z_g,\Upsilon_g$ are the vector of inducing points for the dynamic GP on the  function $g_t$, ${\bf g}=[g_{\delta}\big({\bf u}(\delta)\big),\dots,g_{n\delta}\big({\bf u}(n\delta)\big)]^\top$, and ${\bf u}=[{\bf u}(\delta),\dots,{\bf u}(n\delta)\big)]^\top$.
We have already defined the likelihood term in \eqref{eq:likedeep}.  The second term is
\begin{equation}
\scalebox{0.95}{$
\begin{aligned}
p({\bf f},\Upsilon_f|Z_f,{\bf g})&=N({\bf f};Q_f\Upsilon_f,K_{\boldsymbol{\theta}}({\bf g},{\bf g})-Q_fK_{\boldsymbol{\theta}}(Z_f,Z_f)Q_f^\top),\\
p(\Upsilon_f|Z_f)&=N(\Upsilon_f;{\bf 0},K_{\boldsymbol{\theta}}(Z_f,Z_f)),
\end{aligned}$}
\end{equation}
with $Q_f=K_{\boldsymbol{\theta}}({\bf g},Z_f)K_{\boldsymbol{\theta}}^{-1}(Z_f,Z_f)$. The last term is
\begin{equation}
\scalebox{0.95}{$
\begin{aligned}
p({\bf g},\Upsilon_g|Z_g)&=N({\bf g};Q_g\Upsilon_g,S_{\boldsymbol{\theta}}(T,T)-Q_g S_{\boldsymbol{\theta}}(Z_g,Z_g)Q_g^\top),\\
p(\Upsilon_g|Z_g,{\bf u})&=N(\Upsilon_g;{m}_{\boldsymbol{\theta}}({\bf u}),S_{\boldsymbol{\theta}}(Z_g,Z_g)),
\end{aligned}$}
\end{equation}
with $Q_g=S_{\boldsymbol{\theta}}(T,Z_g)S_{\boldsymbol{\theta}}^{-1}(Z_g,Z_g)$.

We use the following variational posterior approximation:
\begin{equation}
\scalebox{0.95}{$
\begin{aligned}
&q({\bf f},\Upsilon_f,{\bf g},\Upsilon_g)=p({\bf f},\Upsilon_f|Z_f,{\bf g})q(\Upsilon_f)p({\bf g},\Upsilon_g|Z_g,{\bf u})q(\Upsilon_g)
\end{aligned}$}
\end{equation}
where the two variational distributions $q(\Upsilon_f),q(\Upsilon_g)$ are assumed to be Gaussian and their mean and covariance matrix being the variational parameters. We can then derive a lower bound for the evidence as follows:
\begin{equation}
\label{eq:elbo}
\scalebox{0.85}{$
\begin{aligned}
&\log p({\bf y})\\
&\geq \int q({\bf f},\Upsilon_f,{\bf g},\Upsilon_g)\log\tfrac{p({\bf f},\Upsilon_f,{\bf g},\Upsilon_g|Z_f,Z_g,{\bf u})}{q({\bf f},\Upsilon_f,{\bf g},\Upsilon_g)}d{\bf f}d\Upsilon_f d{\bf g} d\Upsilon_g\\
&= \int p({\bf f},\Upsilon_f|Z_f,{\bf g})q(\Upsilon_f)p({\bf g},\Upsilon_g|Z_g,{\bf u})q(\Upsilon_g)\\
&~~~~~~\log\tfrac{p({\bf y}|{\bf f})p(\Upsilon_f|Z_f)p(\Upsilon_g|Z_g)}{q(\Upsilon_f)q(\Upsilon_g)}d{\bf f}d\Upsilon_f d{\bf g} d\Upsilon_g\\
&= \int p({\bf f},\Upsilon_f|Z_f,{\bf g})q(\Upsilon_f)p({\bf g},\Upsilon_g|Z_g,{\bf u})q(\Upsilon_g)\log(p({\bf y}|{\bf f}))d\Upsilon_f d{\bf g} d\Upsilon_g\\
&-KL[q(\Upsilon_f)||p(\Upsilon_f|Z_f)]-KL[q(\Upsilon_g)||p(\Upsilon_g|Z_g,{\bf u})]\\
\end{aligned}$}
\end{equation}
The integral can be efficiently computed using Monte Carlo sampling, as all the densities are Gaussian and they can be further simplified. Furthermore, because the bound factorises over the data, scalability can be achieved through the use of mini-batches. The hyperparameters are estimated by maximising the sum of this evidence lower bound and the logarithm of the prior over the hyperparameters, as in MAP inference.
The derivations for a dynoGP with  a generic architecture is similar. 

\section{Implementation}
\label{app:impl}
We implemented \textit{dynoGP} in \textit{GPyTorch} \cite{gardner2018gpytorch}, and the implementation is available at \url{https://github.com/benavoli/dynoDeepGP}. The main difference compared to a standard DeepGP is that the dynamic GP layer has a covariate time, regardless of its location within the deep network. This also implies that the inducing points for dynamic layers are scalar and represent time. Since time is ordered and the inputs of the dynamic layers are functions of time, it must be carefully managed in the implementation. For this reason, we do not optimise the inducing point locations for dynamic GP layers but only for static layers.
We set the inducing points for the dynamic layer as follows. Let $T = [t_1, t_2, \dots, t_n]$ be the time points of the observations. The inducing points are randomly selected from $T$ without resampling, based on a categorical distribution with unnormalised probabilities 
$p_i \propto \log(1 + t_i / t_n)$. 
This ensures that time instants closer to $t_n$ have a higher probability of being sampled, which improves accuracy in forecasting future values.

We used the following number of inducing points for the dynamic layer and static layer in the numerical experiments:  identification of a simulated Wiener system (800 for the dynamic layer and 200 for the static layer);
 identification of the Wiener-Hammerstein benchmark  (1000 for the dynamic layers and 300 for the static layer); identification the coupled-electric drives benchmark
(250 for the dynamic layers and 175 for the static layer);
 forecast of electricity demand (700 for the dynamic layers and 200 for the static layer).

 The number of training iterations through the dataset and the number of mini-batches per iteration are:
 simulated Wiener system (2000,4);
 identification of the Wiener-Hammerstein benchmark  (1200,8); identification the coupled-electric drives benchmark
(3000,1);
 forecast of electricity demand (500,25).

We use uniform priors over all hyperparameters except for the matrices $L_j$ and $L_j^\dagger$ in the LTI system. For these, we set a horseshoe prior \cite{carvalho2009handling} with a scale parameter of $0.1$. The horseshoe prior is introduced to facilitate model selection between a deterministic and a stochastic LTI system in each of the dynamic GP layers. Its flat, Cauchy-like tails allow $L$ to take large values a posteriori, while its infinitely tall spike at the origin enforces strong shrinkage for elements of $L$ that are close to zero.
\end{document}